\documentclass{article} % For LaTeX2e
\usepackage{iclr2025_conference,times}

\usepackage{amsmath,amsfonts,bm}

\def\eqref#1{equation~\ref{#1}}
\def\1{\bm{1}}

\DeclareMathAlphabet{\mathsfit}{\encodingdefault}{\sfdefault}{m}{sl}
\SetMathAlphabet{\mathsfit}{bold}{\encodingdefault}{\sfdefault}{bx}{n}

\usepackage{hyperref}
\usepackage{url}

\usepackage{mathrsfs}
\usepackage{amsmath}
\usepackage{amsfonts}

\newtheorem{lemma}{Lemma}
\newtheorem{assumption}{Assumption}

\usepackage{multirow}
\usepackage{mathrsfs}
\usepackage{graphics}
\usepackage{subfigure}
\usepackage{wrapfig}
\usepackage{epstopdf}
\usepackage{pdfpages}
\usepackage{diagbox}
\usepackage{ulem}
\usepackage{CJKulem}
\usepackage{pifont} % 包含 \ding 命令
\usepackage{MnSymbol}
\usepackage{xcolor} 
\usepackage{tabularx}
\usepackage{diagbox}
\usepackage{makecell}

\title{Breaking Free from MMI: A New Frontier in Rationalization by Probing Input Utilization}
\author{Wei Liu$^1$,\quad  Zhiying Deng$^2$\footnotemark[1],\quad  Zhongyu Niu$^1$,\quad Jun Wang$^3$\footnotemark[1],\quad Haozhao Wang$^1$\footnotemark[1], \\ \textbf{Zhigang Zeng}$^4$,\quad  \textbf{Ruixuan Li}$^1$\thanks{Corresponding authors.}\\
$^1$School of Computer Science and Technology, HUST\\
$^2$ Faculty of Artificial Intelligence in Education, Central China Normal University\\
$^3$iWudao Tech\quad
$^4$School of Artificial Intelligence and Automation, HUST\\
$^1$\{idc\_lw,zy\_niu, hz\_wang, rxli\}@hust.edu.cn\\
$^2$zhiyingdzy@gmail.com \quad $^3$jwang@iwudao.tech, \quad $^4$zgzeng@hust.edu.cn
}

\iclrfinalcopy % Uncomment for camera-ready version, but NOT for submission.
\begin{document}

\normalem
\maketitle

\begin{abstract}
 Extracting a small subset of crucial rationales from the full input is a key problem in explainability research. The most widely used fundamental criterion for rationale extraction is the maximum mutual information (MMI) criterion. In this paper, we first demonstrate that MMI suffers from diminishing marginal returns. Once part of the rationale has been identified, finding the remaining portions contributes only marginally to increasing the mutual information, making it difficult to use MMI to locate the rest. In contrast to MMI that aims to reproduce the prediction, we seek to identify the parts of the input that the network can actually utilize. 
 {This is achieved by comparing how different rationale candidates match the capability space of the weight matrix.}
The weight matrix of a neural network is typically low-rank, {meaning that the linear combinations of its column vectors can only cover part of the directions in a high-dimensional space (high-dimension: the dimensions of an input vector)}. If an input is fully utilized by the network, {it generally matches these directions (e.g., a portion of a hypersphere)}, resulting in a representation with a high norm. Conversely, {if an input primarily falls outside (orthogonal to) these directions}, its representation norm will approach zero, behaving like noise that the network cannot effectively utilize.  
Building on this, we propose using the norms of rationale candidates as an alternative objective to MMI. 
Through experiments on four text classification datasets and one graph classification dataset using three network architectures (GRUs, BERT, and GCN), we show that our method outperforms MMI and its improved variants in identifying better rationales. 
We also compare our method with a representative LLM (llama-3.1-8b-instruct) and find that our simple method gets comparable results to it and can sometimes even outperform it. Code: \url{https://github.com/jugechengzi/Rationalization-N2R}.
\end{abstract}

\section{Introduction}\label{sec: introduction}
With the success of deep learning, there are growing concerns over the model interpretability. Exploring the theory and technique of interpretable machine learning frameworks is of immense importance in addressing a myriad of issues. For instance, XAI techniques can aid in detecting model discrimination (fairness) \citep{sigmod-debug}, identifying backdoor attacks (security) \citep{li2022backdoor}, and revealing potential failure cases (robustness) \citep{danqi,ecairationale}, among others. Post-hoc explanations, which are trained separately from the prediction process, may not faithfully represent an agent's decision, despite appearing plausible \citep{lipton2016mythos}. In contrast to post-hoc methods, ante-hoc (or self-explaining) techniques typically offer increased transparency \citep{lipton2016mythos} and faithfulness \citep{interlocking}, as the prediction is made based on the explanation itself. There is a stream of research that has exposed the unreliability of post-hoc explanations and called for self-explanatory methods \citep{stopposthoc,falsehope,ren2024}.

\begin{figure}[t]
\centering
    \includegraphics[width=0.7\columnwidth]{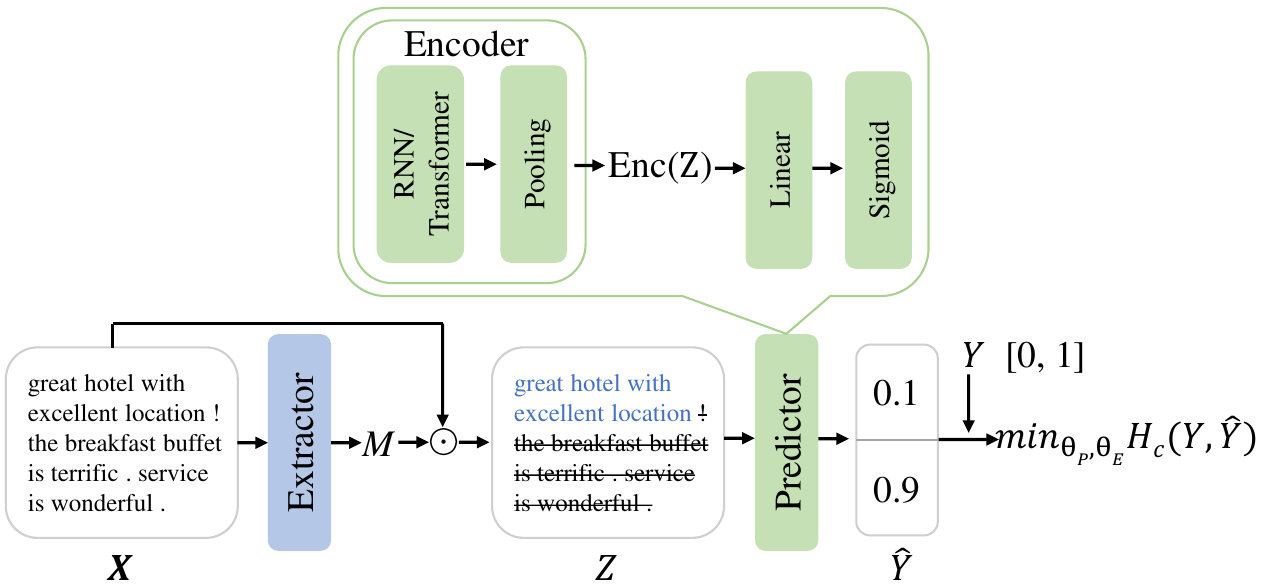}
    \caption{The standard rationalization framework RNP. The task is binary sentiment classification about the hotel's location.  $X,Z,\hat{Y},Y$ represent the input, the extracted rationale candidate, the prediction and the ground truth label, respectively. $M$ is a sequence of binary masks. {$Enc(Z)$ is the encoder's final layer representation (like the term $``$embedding" in \citep{emb1,emb2})}. $\theta_E,\theta_P$ represent the parameters of the extractor and the predictor. $H_c$ denotes  cross-entropy. }
    \label{fig:rnp}
    \vspace{-10pt}
\end{figure}

In this study, our primary focus is on investigating a general model-agnostic self-explaining framework called Rationalizing Neural Predictions (RNP, also known as rationalization) \citep{emnlp/LeiBJ16}, which with its variants has become one of the mainstream methods
to facilitate the interpretability of NLP models \citep{AAAI21learningfrombest,interlocking,aaai-multiaspect,liufr,liumgr,liudr,mcd,dar,mrd,noise,a2i,agr}, and also holds the potential to be applied to image classification \citep{GDM} and graph neural networks \citep{pgexplainer}. RNP utilizes a cooperative game involving an extractor and a predictor. This game is designed with a focus on ``data-centric" (i.e., it is to explain the connection between a text and the model-agnostic task label, rather than explaining the output of a specific model) feature importance. The extractor first identifies the most informative part of the input, termed the rationale. Subsequently, the rationale is transmitted to the predictor to make predictions, as illustrated in Figure~\ref{fig:rnp}.
Apart from its use for interpretability, some recent studies find that rationalization can also serve as a method for data cleaning. The extracted $(Z,Y)$ pairs can act as a new dataset, and trained with such a cleaned dataset, a predictor may be more robust \citep{danqi} and generalizable \citep{dir,gui2023joint}, thanks to the removal of task-irrelevant, harmful information. 

The commonly used objective for finding rationales is maximizing the mutual information between the rationale candidates and the task labels (e.g., by minimizing cross-entropy), which is called the maximum mutual information (MMI) criterion. In practice, however, MMI faces the problem of diminishing marginal utility. For example, if the rationale candidate consists of $80\%$ the real rationales and $20\%$ random noise, it might be informative enough to help a predictor make the correct prediction (see a specific toy example in $\S$\ref{sec: empirical observation}, and empirical verification in Figure \ref{fig:rationale_acc}(a)(b)). In this case, replacing the left  $20\%$ noise with real rationales can only trivially improve the mutual information (or intuitively, the prediction accuracy. See $\S$\ref{sec: empirical observation} for the theoretical perspective). As a result, the gradient provided by MMI cannot guide to find the left $20\%$ real rationales well.

To avoid the shortcomings of MMI, instead of following the traditional approaches of fixing MMI’s flaws through various regularization terms, this paper aims to find an alternative objective to the MMI criterion. 
Since the MMI criterion has been the fundamental objective in the XAI literature for a long time, our finding is important as it opens a new avenue for extractive interpretability without the fundamental MMI. 
The overall idea can be summarized as follows: (1) The rationales are those the predictor can learn and utilize.  
(2) {Neural networks usually have low-rank weight matrices
\citep{lownorm}}, {meaning that the linear combinations of its column vectors can only cover part of the directions in a high-dimensional space (high-dimension: the dimensions of an input vector)}. 
(3) For features that the network does not learn, {their directions are usually orthogonal to the learned directions (see Appendix \ref{app: theoretical support for the low norm}), and the representation norms through the weight matrix will approach zero. On the contrary, well-learned features usually interact with the learned directions of the weight matrix}, resulting in representations with higher norms {(Appendix \ref{app: toy example} provides a toy example for intuitive understanding)}. This property is borrowed from recent advances \citep{lownorm}  in the out-of-distribution (OOD) detection field. 
 Based on the above properties, we attempt to find rationales by maximizing the norm (i.e., $||Enc(Z)||_2$ in Figure~\ref{fig:rnp}) of the rationale candidate's representation, whose motivation is further empirically supported by the results in Figure \ref{fig:rationale_acc}(c).

The key difference between our method and the mainstream MMI approaches can be summarized at a high level: MMI aims to find rationales that can reproduce the prediction, while our study seeks to identify the parts of the input that the network can actually utilize (i.e., match the non-zero rank subspaces of the weight matrix), whose idea is more in line with the philosophy of \textbf{ante-hoc} explanation. 
Our work opens new eyes in the XAI literature, as almost all current mainstream XAI methods follow the MMI criterion or its variants. 
The idea to observe which parts of the input can be utilized by the network from the perspective of forward propagation is also novel, as this is the first time that an extractive XAI method has been freed from relying on the model's final output (e.g., may be used in the future to probe task-irrelevant (i.e., not fine-tuned) pretrained encoders. But this is beyond the scope of this paper and is left for future work.). 

The contributions can be summarized as follows: (1) We empirically find the diminishing marginal utility problem of identifying rationales with the MMI criterion (Figure \ref{fig:rationale_acc}). Considering that MMI is a widely used criterion for finding explanations, this empirical observation may somewhat remind the XAI community to rethink this fundamental criterion.
(2) We formally analyze the reasons why the diminishing marginal utility can occur with the MMI criterion, providing insights for further researchers to better address this issue. (3) Based on a theoretical property borrowed from the OOD research, we propose an alternative objective to the MMI criterion. Empirical results on both text and graph data with three different encoders (GRUs, BERT, and GCN) show that our method not only outperforms the vanilla MMI, but also beats its several recently improved variants.

\section{Related Work}\label{sec: related}
\textbf{Extractive Rationalization}.
The self-explaining framework of rationalization named RNP \citep{emnlp/LeiBJ16} is flexible and offers a unique advantage: certification of exclusion, which means any unselected input is guaranteed to have no contribution to prediction, making it important to the NLP community \citep{interlocking}. Based on it, many methods have been proposed to improve RNP from different aspects. 
\citet{2018rationalegumble} used 
Gumbel-softmax to do the reparameterization for binarized selection. \citet{hardkuma}  replaced the Bernoulli sampling
distributions with rectified Kumaraswamy distributions. \citet{jain2020faith} disconnected the training
regimes of the generator and predictor networks using a saliency threshold. \citet{informationbottle} imposed a discrete bottleneck objective to balance the task performance and the rationale length. \cite{eraser} proposed a benchmark that can be used for supervised rationale extraction.
3PLAYER \citep{rethinking} tries to squeeze the informative texts from the unselected parts to produce comprehensive rationales. DMR \citep{dmr} tries to align the distributions of rationale with the full input text in both the output space and feature space. A2R \citep{interlocking} endows the predictor with the information of full text by introducing a soft rationale. 
FR \citep{liufr} folds the two players to regularize the predictor with the extractor (as the extractor can view the raw input) by sharing a unified encoder.
Inter\_RAT \citep{interventional} tried to use backdoor adjustment to alleviate the spurious correlations in the raw dataset. 
\citet{scott}  leveraged meta-learning techniques to improve the quality of the explanations. \citet{cooperative} cooperatively trained the models with continuous and discrete optimisation schemes. \citep{leakage} explored better metrics for evaluation. \citep{concept} used phrase-based concepts to conduct a self-explaining model. Other methods like data augmentation with pretrained models \citep{counter}, training with human-annotated rationales \citep{Unirex}, injecting noise to the selected rationales \citep{noise}, using attack techniques to inspect spurious correlations \citep{a2i}, have also been tried.

All of these previous studies take MMI as the fundamental criterion of finding rationales, and the diminishing marginal utility problem has been overlooked. The purpose of this paper is to analyze the  diminishing marginal utility problem and to alleviate it, which is orthogonal to previous research.

\textbf{The properties of the network's learned inputs}. 
Some previous research has found that complex neural networks typically have low-rank weight matrices \citep{AghajanyanGZ20}. \cite{lownorm} shows both theoretically and empirically that the weight matrices and network representations associated with the learned inputs often occupy low-dimensional subspaces with high overlap. However, when the network encounters unlearned OOD inputs, their associated representations tend to have less overlap with the weight matrices compared to those the network has learned. As a result, the feature representations corresponding to the unlearned OOD inputs tend to have smaller norms than those of the learned inputs, resulting in less signal being propagated from the input. Our method is inspired by this theoretical property.

\textbf{Generative explanation with LLMs}.
Generative explanation is a research line that is close but orthogonal to our research on extractive explanation. With the great success of LLMs, a new research line for explanation is chain-of-thought. By generating (in contrast to selecting) intermediate
reasoning steps before inferring the answer, the reasoning steps can be seen as a kind of explanation. The intriguing
technique is called chain-of-thought (CoT) reasoning \citep{cot}. However, LLMs sometimes exhibit unpredictable failure modes \citep{causalllm} or hallucination reasoning \citep{surveyofllm}, making this kind of generative explanation not trustworthy enough in some high-stakes scenarios. Also, some recent research finds that LLMs are not good at extractive tasks \citep{chatgptgeneral,evaluatingchatgpt,comprehensivechatgpt}.

\textbf{The potential impact of rationalization in the era of LLMs}.
Compared to traditional ``model-centric" XAI methods which solely focus on the model's learned information, ``data-centric" approaches primarily aim to extract model-agnostic patterns inherent in the data. So, apart from improving interpretability, rationalization can serve as a method of data cleaning \cite{datacentric}.

Domain-specific large models often require supervised fine-tuning using domain-specific data. Uncleaned data may contain harmful information such as biases and stereotypes \citep{trustllm}. Recent research suggests that training predictors with extracted rationales can remove irrelevant harmful information, enhancing robustness \citep{danqi} and generalization \citep{dir,gui2023joint}. Considering that small models are sufficient for simple supervised tasks and are more flexible and cost-effective for training on single datasets (e.g., searching hyperparameters and adding auxiliary regularizers), using small models for rationalization on a single dataset and then using the extracted rationales for supervised fine-tuning might prevent large models from learning harmful information from new data. Additionally, shortening input texts can also reduce the memory required for fine-tuning \citep{skimformer}. A recent study also finds that training a small model for data selection (although not the same as rationale selection) and producing a small subset is useful for fine-tuning LLMs \citep{less}.

We compare our method against a representative LLM (llama-3.1-8b-instruct), in Appendix \ref{app: llm result}, and demonstrate that our approach achieves comparable results, sometimes even surpassing it.

\section{Preliminaries}
\subsection{The rationale extraction task}\label{sec: rationalization task}
We consider the text classification task, where the input is a text sequence  ${X}$=$[x_1,x_2,\cdots,x_l]$ with ${x}_i$ being the $i$-th token and $l$ being the number of tokens. $Y$ represents the classes in a dataset $\mathcal{D}$. The standard rationalization framework RNP \citep{emnlp/LeiBJ16} consists of an extractor $f_E(\cdot)$ and a predictor $f_P(\cdot)$, with $\theta_e$ and $\theta_p$ representing the parameters of the extractor and predictor. 
For  $(X,Y)\sim\mathcal{D}$, the extractor first outputs a sequence of binary mask $M=f_E(X)=[m_1,\cdots,m_l]\in \{0,1\}^l$ (in practice, the extractor first outputs a Bernoulli distribution for each token and the mask for each token is independently sampled using gumbel-softmax). Then, it forms the rationale candidate $Z$ by the element-wise product of $X$ and $M$:
\begin{equation}\label{eqa:getrat}
    Z=M\odot X=[m_1x_1,\cdots,m_lx_l].
\end{equation}
To simplify the notation, we denote $f_E(X)$ as $Z$ in the following sections, i.e., $f_E(X)=Z$. With the extractor's selection, we get a set of $(Z,Y)$ samples, which are generally considered to represent the distribution $P(Y|Z)$. The rationale $Z$ is searched by maximizing the mutual information $I(Y;Z)$:

\begin{equation}\label{eqa: mmi}
\begin{aligned}
      Z^*&=\mathop{\arg\max}_{Z}I(Y;Z)
      =\mathop{\arg\max}_{Z}(H(Y)-H(Y|Z))=\mathop{\arg\min}_{Z}H(Y|Z), \ s.t., \ Z=f_E(X).  
\end{aligned}
\end{equation}

In practice, the entropy $H(Y|Z)$ is commonly approximated by the minimum cross-entropy $\mathop{\min}_{\theta_p}H_c(Y,\hat{Y}|Z)$, with $\hat{Y}=f_P(Z)$ representing the output of the predictor. 
It is essential to note that the minimum cross-entropy is equal to the entropy (please refer to Appendix~\ref{app: minimizing cross entropy is equal to kl}).

As a result, the predictor uses the cross-entropy objective to do the classification, and the extractor also uses the cross-entropy objective to find good rationales:

\begin{equation}\label{eqa: disentangle rnp}
\begin{split}
   &\textbf{Extractor:}  \   \mathop{\min}_{\theta_e}H_c(Y,f_P(Z)|Z)
\\
    &\textbf{Predictor:} \ \mathop{\min}_{\theta_p}H_c(Y,f_P(Z)|Z)
\\
&s.t., \ Z=f_E(X), \ (X,Y)\sim \mathcal{D}.
\end{split}
\end{equation}

Replacing $Z$ with $f_E(X)$, the extractor and the predictor are trained cooperatively to minimize the cross-entropy. Here we rewrite Equation (\ref{eqa: disentangle rnp}) for better conciseness and clarity: 
\begin{equation}\label{eqa:overall prediction accuracy}
\mathop{\min}_{\theta_e, \theta_p}H_c(Y,f_P(f_E(X))|f_E(X)), \ s.t., \ (X,Y)\sim \mathcal{D}.
\end{equation}

To make the selected rationale human-intelligible, rationalization methods usually constrain the rationales by compact and coherent regularization terms. In this paper, we use the most widely used constraints proposed by  \cite{invarant}:
\begin{equation}\label{eqa:sparse regular}
\Omega (M) = \lambda_1 \bigg \lvert \frac{||M||_1}{l}-s \bigg\rvert +\lambda_2\sum_{t=2}^{l} \big|m_t-m_{t-1} \big|. 
\end{equation} The first term encourages that the percentage of the tokens being selected as rationales is close to a pre-defined level $s$. The second term encourages the rationales to be coherent.

\subsection{The properties of the network's utilization on different inputs}\label{sec: ood property}
It has been found that neural networks usually have low-rank weight matrices \citep{lownorm}. For these low-rank weight matrices, they correspond to low-dimensional subspaces. And the location and magnitude of these subspaces are determined by the information they learn \citep{lownorm}. If a network just learns (or is trained on) some uninformative noise, the rank will be low and the corresponding subspace will be narrow. On the contrary, if the network learns from informative knowledge, the subspace will be wider. From a localization perspective, the learned inputs often occupy these low-dimensional subspaces with high overlap. And the unlearned inputs tend to have little overlap. As a result, if an input is informative and the knowledge is learned by the network, the representation of it through the network will have a high ${l}_2$ norm. If an input does not contain the knowledge learned by the network, the norm will approach $0$ \citep{lownorm}. 

Simply put, the rank of the weight matrix is determined by the knowledge learned by the network and {is reflected in the directions on the hypersphere that can be occupied by the combination of the column vectors (including position and size). These areas occupy a subspace in the high-dimensional space, which we refer to as the capability subspace of the weight matrix}. If an input contains features that the network has learned, {it is highly likely to fall within the capability subspace and match the learned directions}, leading to a representation with a higher norm. On the other hand, {if an input does not contain the learned information, it is likely to fall outside the capability subspace and be orthogonal to the column vectors (see Appendix \ref{app: theoretical support for the low norm})}, thus its representation norm will be very low, behaving like noise. At the same time, if the network does not learn any informative knowledge, the capability subspace of the weight matrix itself will be very low and all inputs will have low norms. 

This property was first used by \cite{lownorm} to remove out-of-distribution inputs in reinforcement learning. But we think it can also be used for interpretability. By observing the norms of the representations of different rationale candidates, we can determine how well they match the network, and thus identify which parts of the full input the network is actually utilizing.

\section{The limitations of minimizing cross-entropy and our method}\label{sec: limitations of mmi}
\subsection{The diminishing marginal returns}\label{sec: empirical observation}
Although using MMI to identify rationales has almost become the default choice, we find that it faces the problem of diminishing marginal returns. Once the majority of the rationale components have been identified, discovering the left rationale components has a minimal effect on further reducing cross-entropy.

In this section, we first provide an intuitive toy example to help readers understand the diminishing marginal returns from a high-level intuition. 
Then, we show the origin of the diminishing marginal returns problem. Finally, we provide empirical evidence from real-world datasets to verify the existence of this problem in practice.

\begin{wrapfigure}[15]{R}{0.4\columnwidth}
    % \vspace{-10pt}
    \includegraphics[width=0.4\columnwidth]{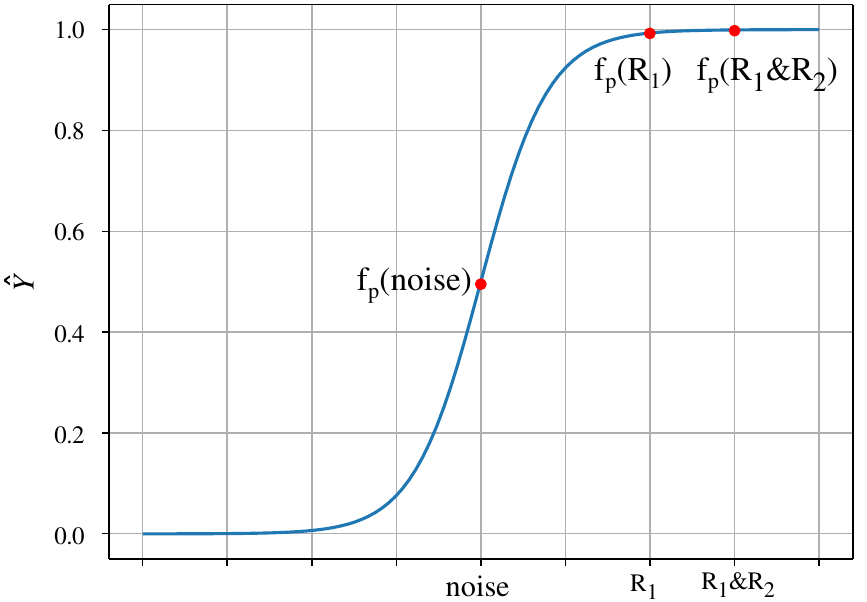}
    % \vspace{-10pt}
  \caption{The diminishing marginal returns in Sigmoid function.}
  \label{fig: sigmoid}
\end{wrapfigure}

\textbf{An intuitive toy example}. Consider a comment about food ``$\cdots\cdots$ The food is very delicious, and I like it very much. $\cdots\cdots$'' and we need to predict its sentiment label. Both $R_1=$``The food is very delicious'' and $R_2=$``and I like it very much'' can indicate a positive enough sentiment tendency. However, if either $R_1$ or $R_2$ is given, finding another component will not contribute much to the sentiment polarity . This creates an obstacle to finding the complete rationale by MMI (minimizing the cross-entropy in practice). A more visual example is provided in Figure \ref{fig: sigmoid}. The distance between $R_1$ and $R_1\&R_2$ is non-trivial, so the extractor needs to pay considerable effort to move from the state of selecting $R_1$ to the state of selecting both $R_1$ and $R_2$. However, the payoff for this effort is minimal, resulting in small gradients provided by the gradient descent algorithm for this move.

The reasons of this problem can lie in two aspects. One is the gradient saturation problem of the sigmoid function before the predictor's output $\hat{Y}$, which is quite intuitive and can be illustrated by the example in Figure \ref{fig: sigmoid} (softmax is similar). Aside from the sigmoid function, the problem can also be introduced by the diminishing marginal returns of mutual information itself. 

\begin{figure*}[t]
    % \flushleft
      \subfigure[]{
            \includegraphics[width=0.31\columnwidth]{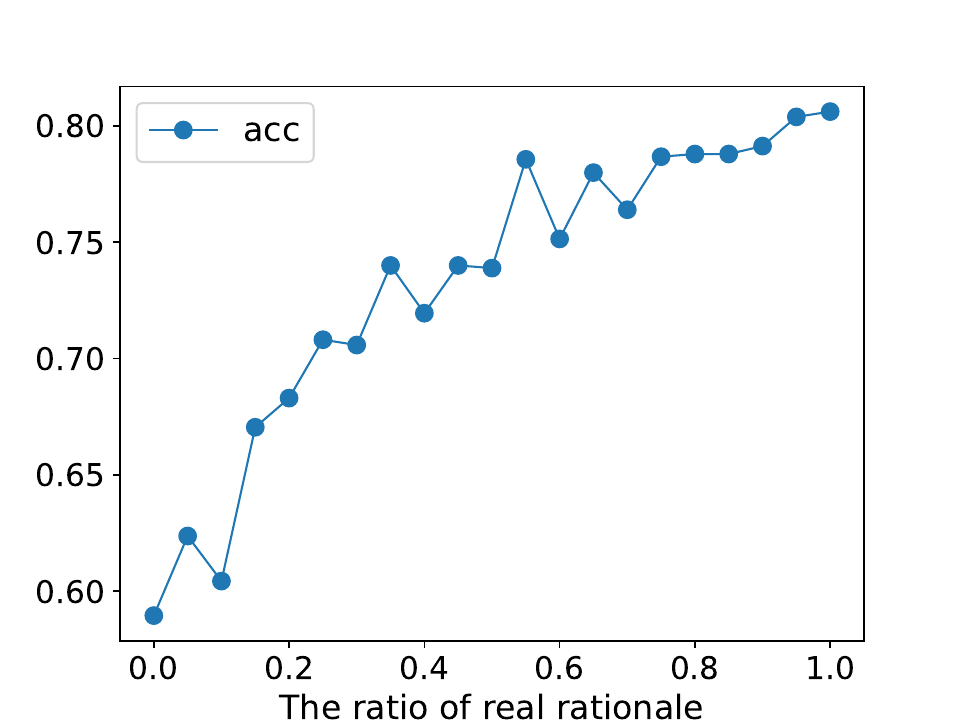}
        \label{fig:rationale_acc1}
        }
        \subfigure[]{
            \includegraphics[width=0.31\columnwidth]{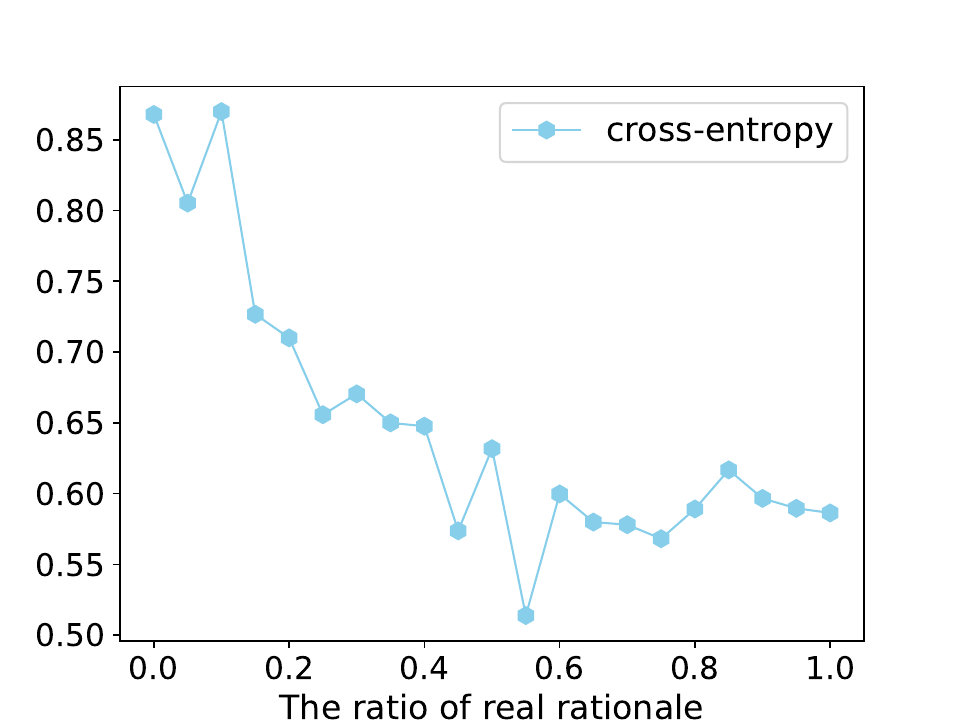}
            \label{fig:rationale_loss}
        }
        \subfigure[]{
            \includegraphics[width=0.31\columnwidth]{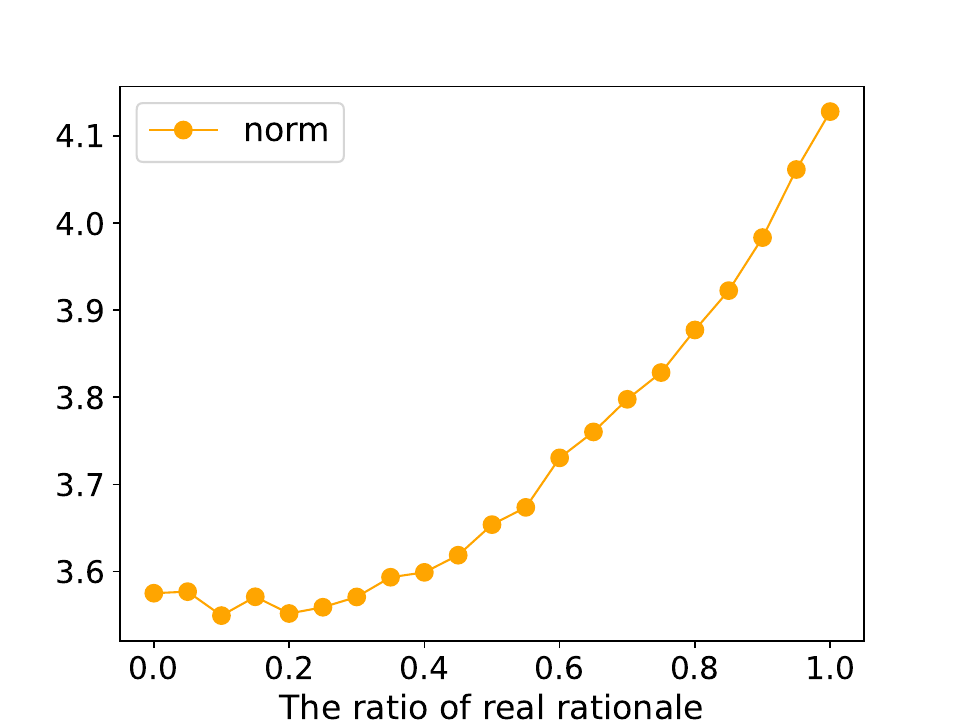}
            \label{fig:rationale_norm}
        }

    \caption{The (a) prediction accuracy, (b) cross-entropy loss, and (c) the norm of the representation (i.e., $||Enc(Z)||_2$) through the neural network vary with the proportion of true rationale components in the rationale candidate input within a trained standard RNP predictor. The dataset is \emph{Beer-Aroma}. The results of more datasets are shown in Appendix \ref{app: marginal return}.}
    \label{fig:rationale_acc}
    % \vspace{-10pt}
\end{figure*}

\textbf{The theoretical perspective}. The high-level understanding of mutual information $I(Y;Z)$ is that, how much the uncertainty of $Y$ decreases when given $Z$.

We consider a perfect rationale $R$ composed of $R_1$ and $R_2$. The problem is that MMI does not satisfy additivity, which means that we cannot promise $I(Y;R_1,R_2)= I(Y;R_1)+I(Y;R_2)$. 
The combined effect of $R_1$ and $R_2$ on reducing the uncertainty of $Y$ may be less than the sum of the individual effects of $R_1$ and $R_2$ on reducing the uncertainty of $Y$. Formally, we consider the situations where 
\begin{equation}\label{eqa: mmi with two variable}
    I(Y;R_1,R_2)\leq I(Y;R_1)+I(Y;R_2).
\end{equation}
Please refer to Appendix \ref{app: proof of mmi with two variable} for more detailed  discussions about Equation (\ref{eqa: mmi with two variable}).

From Equation (\ref{eqa: mmi with two variable}), we can further get 
\begin{equation}
\begin{aligned}
    I(Y;R_2|R_1)&=I(Y;R_2,R_1)-I(Y;R_1) \\
    &\leq [I(Y;R_1)+I(Y;R_2)]-I(Y;R_1)\\
    &=I(Y;R_2).
\end{aligned}
\end{equation}
It means that, although $R_2$ is informative and can reduce the uncertainty of $Y$, it can be less effective if conditioned on another informative part $R_1$.

\textbf{Empirical verification of the diminishing marginal returns}.  
Figure \ref{fig:rationale_acc} shows how the prediction accuracy, cross-entropy loss, and the norm of the representation (i.e., $||Enc(Z)||_2$ in Figure \ref{fig:rnp}) vary with the proportion of true rationale components in the rationale candidate input within a trained standard RNP predictor. The dataset is a text classification dataset \emph{Beer-Aroma}. The results of more datasets are shown in Appendix \ref{app: marginal return}.

The test set contains human-annotated ground-truth rationales. 
We base our inputs on the manually annotated ground-truth gold rationales in the original full text. A certain proportion of tokens in the gold rationales are replaced with tokens randomly selected from the non-rationale part of the same full text. The x-axis represents the proportion of tokens that have not been replaced, where 0 indicates all tokens are random, and 1 indicates the complete ground-truth rationale. It can be observed that when the rationale candidate used as input contains more than $60\%$ true rationale components, the decrease in cross-entropy loss (Figure \ref{fig:rationale_acc}(b)) and the increase in accuracy (Figure \ref{fig:rationale_acc}(a)) slow down. Subsequent addition of new rationale components has a diminished effect on loss reduction, making it difficult to identify the left rationale components. 

This raises a question: is there an objective function that can indicate rationale, and does not involve either the mutual information or the sigmoid function of the neural network's output layer, thereby allowing for the possibility of avoiding diminishing marginal returns?

\textbf{Empirical observation on the norm of intermediate representation}.
From the theoretical property mentioned in $\S$\ref{sec: ood property}, we know that the representation's norm is an indicator for the degree of the network's utilization on an input. Besides, it does not involve either mutual information or the sigmoid function. We then empirically observe whether it faces the problem of diminishing marginal returns. Figure \ref{fig:rationale_acc}(c) shows how the norm varies with the proportion of true rationale components in the rationale candidate input within a trained standard RNP predictor. We see that it indeed does not face the problem of diminishing marginal returns. Instead, it grows even faster as the proportion of rationale components grows. 

\subsection{The practical method}

Unlike existing mainstream rationalization methods that add extra auxiliary modules, we do not change the model architecture of the vanilla RNP, thus preserving its conciseness and flexibility. The only needed modification is to replace the extractor's cross-entropy loss with the norm loss.

Compared to MMI-based methods (Equation \ref{eqa: disentangle rnp}), we remove the cross-entropy loss from the extractor's parameters $\theta_e$ and replace it with the norm:
\begin{align}
\begin{split}\label{eqa: extractor objective}
   &\textbf{Extractor:}  \   \mathop{\min}_{\theta_e} -log (||Enc(Z)||_2) 
\end{split}
\\
\begin{split}\label{eqa: predictor objective}
    &\textbf{Predictor:} \ \mathop{\min}_{\theta_p}H_c(Y,f_P(Z)|Z)
\end{split}
\\
&s.t., \ Z=f_E(X), \ (X,Y)\sim \mathcal{D}.
\end{align}
The predictor is trained to do the classification, and the extractor is trained to identify the rationales. During training, (\ref{eqa: predictor objective}) and (\ref{eqa: extractor objective}) are alternated.
The practical implementation with Pytorch is in Appendix \ref{app: implementation of N2R}.
We call this method N2R (norm to rationale). 
Note that when we say MMI (see Equation (\ref{eqa: mmi})), it refers only to the objective of the extractor , and does not include the predictor. No matter how the extractor's objective changes, the predictor is always trained with the cross-entropy as it needs to do the classification.

To further show the potential and scalability of our N2R, we also verify the possibility of combining MMI and N2R (corresponding to the supplementary experiments in Figure \ref{fig: noacc}). This combination is inspired by the empirical results of Figure \ref{fig:rationale_acc}. 
In the initial stage of training, the extractor has not yet identified enough true rationale components, leading to small gradients provided by the norm objective (Figure \ref{fig:rationale_acc}(c)), which is not efficient enough to guide the extractor in finding the rationale. However, at this stage, the MMI objective provides larger gradients (Figure \ref{fig:rationale_acc}(b)). In the later stages of training, the situation is reversed. Overall, the MMI and norm objectives complement each other, resulting in improved performance.
We maintain the objective of the vanilla RNP (Equation \ref{eqa: disentangle rnp}) and add the norm to the extractor's objective:
\begin{align}
\begin{split}
   &\textbf{Extractor:}  \   \mathop{\min}_{\theta_e} [H_c(Y,f_P(Z)|Z)-log (||Enc(Z)||_2) ]
\end{split}
\\
\begin{split}
    &\textbf{Predictor:} \ \mathop{\min}_{\theta_p}H_c(Y,f_P(Z)|Z)
\end{split}
\\
&s.t., \ Z=f_E(X), \ (X,Y)\sim \mathcal{D}.
\end{align}
We call this method MMI+N2R. The practical implementation with Pytorch is in Appendix \ref{app: implementation of N2R+RNP}.

\section{Experiments}
\subsection{Settings}
\textbf{Baselines}.
The main baseline for direct comparison is the vanilla MMI-based rationalization framework RNP \citep{emnlp/LeiBJ16}, as RNP and our N2R match in selection granularity, optimization algorithm, and model architecture, which helps us to focus on our claims rather than some potentially unknown mechanisms. 
To show the competitiveness of our method, we also include several recently published methods that improve MMI with various regularizers: Inter\_RAT \citep{interventional}, NIR \citep{noise}, CR \citep{cr}, and A2I \citep{a2i}, all of which have been discussed in $\S$\ref{sec: related}. 

\textbf{Datasets}. Although the rationale extraction process is unsupervised, the rationalization task requires comparing the rationale quality extracted by different models. This necessitates that the test set includes ground-truth rationales, which imposes special requirements on the datasets.
Following the conventional setup in the field of rationalization, we employ four text classification datasets from two widely used benchmarks. Apart from the text data, we also include a graph classification dataset. 

The text classification datasets are \textbf{Beer-Appearance, Beer-Aroma} (collected from the BeerAdvocate benchmark \citep{beer}), \textbf{Hotel-Service, Hotel-Cleanliness} (collected from the HotelReviews benchmark \citep{hotel}). We also use a graph classification dataset, called  \textbf{BA2Motifs} \citep{gnnexplainer}, to verify generalizability. All of these datasets contain human-annotated ground-truth rationales on the test set, making it convenient to compare different methods' performance fairly. More details are in Appendix \ref{app: datasets}.

\textbf{Implementation details}. Both  the extractor and the predictor  are composed of an encoder (RNN/Transformer/GCN) and a linear layer. We use three types of encoders: GRUs (following Inter\_RAT and A2I, table \ref{tab: beer} and \ref{tab: hotel}), bert-base-uncased (following CR, table \ref{tab: bert results}), and GCN (for the BA2Motifs dataset). For NIR and our N2R, considering they are both variants of the standard RNP, we first manually tune the hyperparameters for RNP, and then apply the hyperparameters to both NIR and N2R. For Inter\_RAT, since it has originally been implemented on the beer-related datasets, we apply its original hyperparameters but only adjust the sparsity regularizer in Equation (\ref{eqa:sparse regular}). For CR, we just keep the major settings (``bert-base-uncased", the Beer-Appearance dataset, and the sprasity of $10\%$) the same as it and copy its results from its original paper.
We report the average results of five random seeds. More details are in Appendix \ref{app: experimental details}.

\begin{table*}[t]
    \centering
       \caption{Results on datasets from the BeerAdvocate Benchmark. We report the average results of five random seeds. Values in ``()" are the standard deviations. Inter\_RAT: \citet{interventional}. NIR: \citet{noise}. A2I: \cite{a2i}.}
       \vskip 0.1in
       {\resizebox{0.99\columnwidth}{!}{
       \setlength\tabcolsep{4pt}
    \begin{tabular}{c c c |c c| c c c|c c| c c c }
\hline
\multicolumn{3}{c|}{\multirow{2}{*}{\diagbox{Methods}{Datasets}}} & \multicolumn{5}{c|}{Beer-Appearance} & \multicolumn{5}{c}{Beer-Aroma} \\
\cline{4-13}
\multicolumn{3}{c|}{} &S& Acc & P & R &\multicolumn{1}{c|}{F1} &S& Acc & P & R &\multicolumn{1}{c}{F1} \\
\hline

\multicolumn{3}{c|}{RNP} & 14.7 (0.7)& 78.2 (3.3) & 75.0 (0.5) & 59.7 (3.1) & 66.5 (2.1)& 15.2 (1.0) & 81.7 (2.4) & 67.0 (12.1) & 64.7 (8.8)&65.8 (10.4)\\
\multicolumn{3}{c|}{Inter\_RAT} & 15.2 (1.1) & N/A & 57.0 (5.3) & 46.9 (2.3) &51.4 (3.2) & 16.1 (0.7) &N/A& 57.9 (2.4) & 60.3 (2.5) & 59.0 (2.1) \\
% \multicolumn{3}{c|}{NIR} & 15.5 (0.4)& 80.8 (1.2)  &73.3 (1.3) & 61.3 (2.2) & 66.8 (1.7) & 15.2 (0.2) & 82.1 (7.1) & 69.8 (6.0)&67.9 (5.3) & 68.8 (5.6)  \\
\multicolumn{3}{c|}{NIR} & 14.8 (0.4)& 78.2 (2.2)  &74.0 (1.3) & 59.0 (2.4) & 65.6 (2.0) &  15.4 (0.4) & 82.2 (3.2) & 65.4 (7.1) & 64.7 (6.2) & 65.1 (6.6)\\
\multicolumn{3}{c|}{A2I} & 14.9 (0.3) & 81.0 (1.2) & 75.2 (0.9) & 60.6 (1.7) & 67.1 (1.3) & 14.8 (0.1) & 82.7 (2.3) & 69.4 (2.5) & 65.9 (2.6) & 67.6 (2.5)\\
\multicolumn{3}{c|}{N2R (ours)} & 14.8 (0.5) & \textbf{82.3} (1.8) & \textbf{81.9} (2.7) & \textbf{65.3} (2.2) & \textbf{72.7} (2.1) & 14.9 (0.4) & \textbf{86.9} (4.5) & \textbf{70.2} (1.5) & \textbf{67.2} (1.3) & \textbf{68.7} (1.1)\\

\hline
\end{tabular}
}

} 
    \label{tab: beer}

\end{table*}

\begin{table*}[t]
    \centering
       \caption{Results on datasets from the HotelReviews Benchmark. }
       \vskip 0.1in
       {\resizebox{0.99\columnwidth}{!}{
       \setlength\tabcolsep{4pt}
    \begin{tabular}{c c c |c c| c c c|c c| c c c }
\hline
\multicolumn{3}{c|}{\multirow{2}{*}{\diagbox{Methods}{Datasets}}} & \multicolumn{5}{c|}{Hotel-Service} & \multicolumn{5}{c}{Hotel-Cleanliness} \\
\cline{4-13}
\multicolumn{3}{c|}{} &S& Acc & P & R &\multicolumn{1}{c|}{F1} &S& Acc & P & R &\multicolumn{1}{c}{F1} \\
\hline

\multicolumn{3}{c|}{RNP} &  15.3 (0.3) &96.5 (1.5) & 41.0 (1.5) & 54.6 (1.1) &46.8 (1.4) & 15.3 (0.2) & 97.2 (1.6) & 28.1 (0.7) & 48.7 (1.1)&35.6 (0.8) 
\\
\multicolumn{3}{c|}{Inter\_RAT}  & 15.0 (0.8) & N/A& 28.9 (1.1) & 38.1 (1.9) & 32.8 (1.1) & 14.4 (1.1)&N/A & 27.2 (2.1) & 44.1 (2.4) & 33.6 (2.1)
\\
\multicolumn{3}{c|}{NIR} &  15.0 (0.3) & 96.9 (0.4) & 40.9 (1.5) & 53.5 (1.2) & 46.3 (1.4) & 15.5 (0.4) & 96.7 (0.9)& 28.0 (0.6) & 49.2 (1.0) & 35.7 (0.6)
\\
\multicolumn{3}{c|}{A2I} &  15.1 (0.4) & 96.7 (0.6) & 41.4 (1.7) & 54.6 (1.3) & 47.1 (1.5) & 15.3 (0.3) & 96.8 (1.0) & 28.8 (0.6) & 49.7 (1.1) & 36.5 (0.7)
\\
\multicolumn{3}{c|}{N2R (ours)} & 15.1 (0.2) & \textbf{97.4} (0.5) & \textbf{42.8} (0.5) & \textbf{56.3} (1.0) & \textbf{48.6} (0.6) & 14.8 (0.2) & \textbf{97.4} (0.4) & \textbf{31.8} (0.5) & \textbf{53.4} (0.8) & \textbf{39.8} (0.6)\\

\hline
\end{tabular}
}

} 
    \label{tab: hotel}

\end{table*}

\textbf{Metrics}. Following the previous research of Inter\_RAT and A2I, we mainly focus on the rationale quality, which is measured by the overlap between the human-annotated rationales and the model-selected tokens. The terms $P, R, F1$ denote precision, recall, and $F1$ score respectively. 
These metrics are the most frequently used in rationalization. The term $S$ represents the average sparsity of the selected rationales, that is, the percentage of selected tokens in relation to the full text. Since the sparsity of ground-truth rationales on these datasets is around $10\%\sim20\%$, we adjust $s$ in Equation (\ref{eqa:sparse regular}) to make $S$ be about $15\%$ (since Equation (\ref{eqa:sparse regular}) is only a soft constraint, it cannot strictly limit $S$ to be exactly $15\%$.).
$Acc$ stands for the predictive accuracy.

\subsection{Results}
\textbf{Results on standard benchmarks}. Tables \ref{tab: beer} and \ref{tab: hotel} show the results on the four text classification datasets. In terms of the rationale quality (F1 score), our N2R significantly outperforms the standard MMI-based method (i.e., RNP) and also beats its improved variants.
Compared to the second-best results of previous methods, the relevant improvements of our N2R on these four datasets are $8.3\%$ ($=\frac{72.7-67.1}{67.1}$), $1.6\%$ ($=\frac{68.7-67.6}{68.8}$), $5.1\%$ ($=\frac{49.5-47.1}{47.1}$), and $9.0\%$ ($=\frac{39.8-36.5}{36.5}$), showing the competitiveness of replacing the MMI-based objective with our norm objective.

We also compare with a representative LLM, llama-3.1-8b-instruct, in Table \ref{tab: results of llm} of Appendix \ref{app: llm result}, and find that our simple N2R gets comparable results to it and can sometimes even outperform it.

\textbf{Results with BERT encoder}. We also follow CR to conduct experiments with pretrained bert-base-uncased as a supplement. Since some methods become highly sensitive to hyperparameters after switching to an over-parameterized BERT model (also supported by Remark 6.1 in \citep{cr}), and our computational resources are insufficient for extensive hyperparameter tuning for these methods, we primarily compare our approach with methods that have already been implemented using BERT. The results are shown in Table \ref{tab: bert results}. Our N2R still outperforms previous MMI-based methods significantly.

\begin{table}[t]
    \centering
    \caption{Results with BERT encoder.  $``*"$: the results of baselines are obtained from the paper of CR \citep{cr}. }
    
    \vskip 0.1in
    \resizebox{0.99\columnwidth}{!}{
% \cline{4-8}
\setlength\tabcolsep{4pt}
\begin{tabular}{c c c |c c| c c c|c c| c c c }
\hline
\multicolumn{3}{c|}{\multirow{2}{*}{\diagbox{Methods}{Datasets}}} & \multicolumn{5}{c|}{Beer-Appearance} & \multicolumn{5}{c}{{Beer-Aroma}} \\
\cline{4-13}
\multicolumn{3}{c|}{} &S& Acc & P & R &\multicolumn{1}{c|}{F1} &  {S}& {Acc} & {P} & {R} &\multicolumn{1}{c}{  {F1}} \\
\hline
         \multicolumn{3}{c|}{RNP*}&10.0 (n/a)& {91.5 (1.7)} & 40.0 (1.4) & 20.3 (1.9) & 25.2 (1.7) & {10.0 (n/a)} & {84.0 (2.1)} & {49.1 (3.2)} & {28.7 (2.2)} & {32.0 (2.5)} \\
         \multicolumn{3}{c|}{A2R*} & 10.0 (n/a)&  {91.5 (2.2)} & 55.0 (0.8)& 25.8 (1.6) & 34.3 (1.4)& {10.0 (n/a)} & {85.5 (1.9)}& {61.3 (2.8) }& {34.8 (3.1)} & {41.2 (3.3)} \\
         \multicolumn{3}{c|}{INVRAT*} &10.0 (n/a) & {91.0 (3.1)}& 56.4 (2.5) & 27.3 (1.2) & 36.7 (2.1)& {10.0 (n/a)} & {90.0 (3.0)} & {49.6 (3.1)} & {27.5 (1.9)} &  { 33.2 (2.6)}\\
         \multicolumn{3}{c|}{CR*}&10.0 (n/a)& {92.4 (1.7)}& 59.7 (1.9)&31.6 (1.6)&39.0 (1.5)& {10.0 (n/a)} & {86.5 (2.1)} & {68.0 (2.9)} & {42.0 (3.0)} & {49.1 (2.8)}\\
         \multicolumn{3}{c|}{N2R (ours)}& 10.8 (0.3)&  {\textbf{93.5} (1.8)} & \textbf{79.7} (4.1) & \textbf{36.3} (1.8) & \textbf{49.9} (2.5)& {10.0 (0.1)} & {\textbf{91.0} (3.6)} & {\textbf{74.3} (5.8)} & {\textbf{47.0} (3.7) }& {\textbf{57.6} (4.5)}\\
         \hline

    \end{tabular}
   }
   
    \label{tab: bert results}

\end{table}

\textbf{Results with GCN encoder}. To show generalizability of our method, we expand the RNP framework to graph neural networks to conduct a supplement experiment. Since Inter\_RAT, NIR, and CR are methods specifically designed for text data and are not suitable for graph tasks, we only compare our N2R with the standard RNP on the BA2Motifs dataset to show its effectiveness rather than competitiveness. For this dataset, we select a set of nodes for each graph as the rationale.  The results are shown in Table \ref{tab: ba2motif}. We see that our method is still effective when applied to graph neural networks. Note that our method is very simple and has the potential to be combined with more advanced methods in the future. However, since the interpretability of graph neural networks is not the focus of this paper, we leave it for future work.

\begin{figure}[t]
    % \flushleft
    \centering
    \subfigure[]{
\includegraphics[width=0.48\columnwidth]{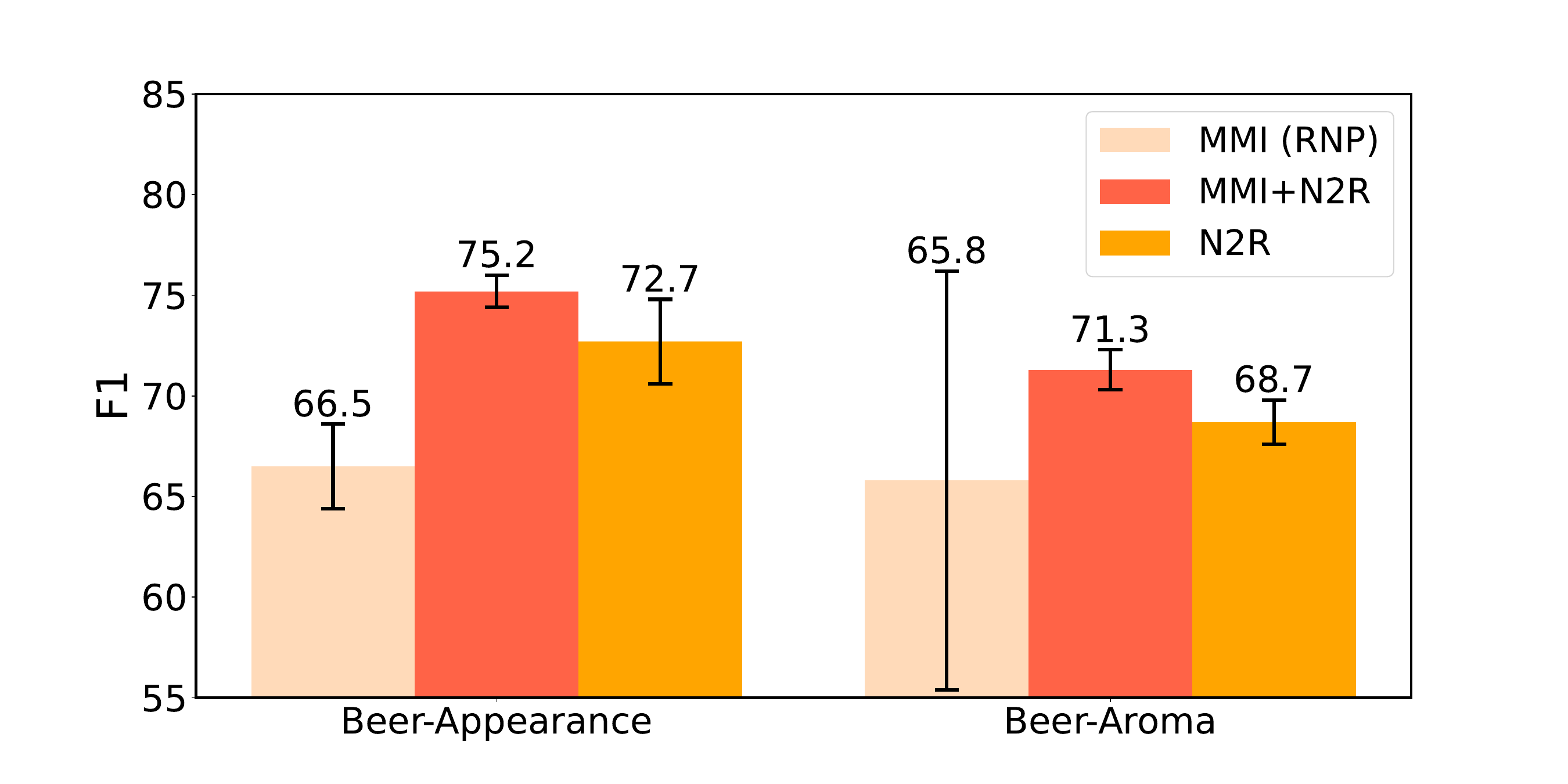}
    \label{fig: beer noacc}
        }
   \subfigure[]{
\includegraphics[width=0.48\columnwidth]{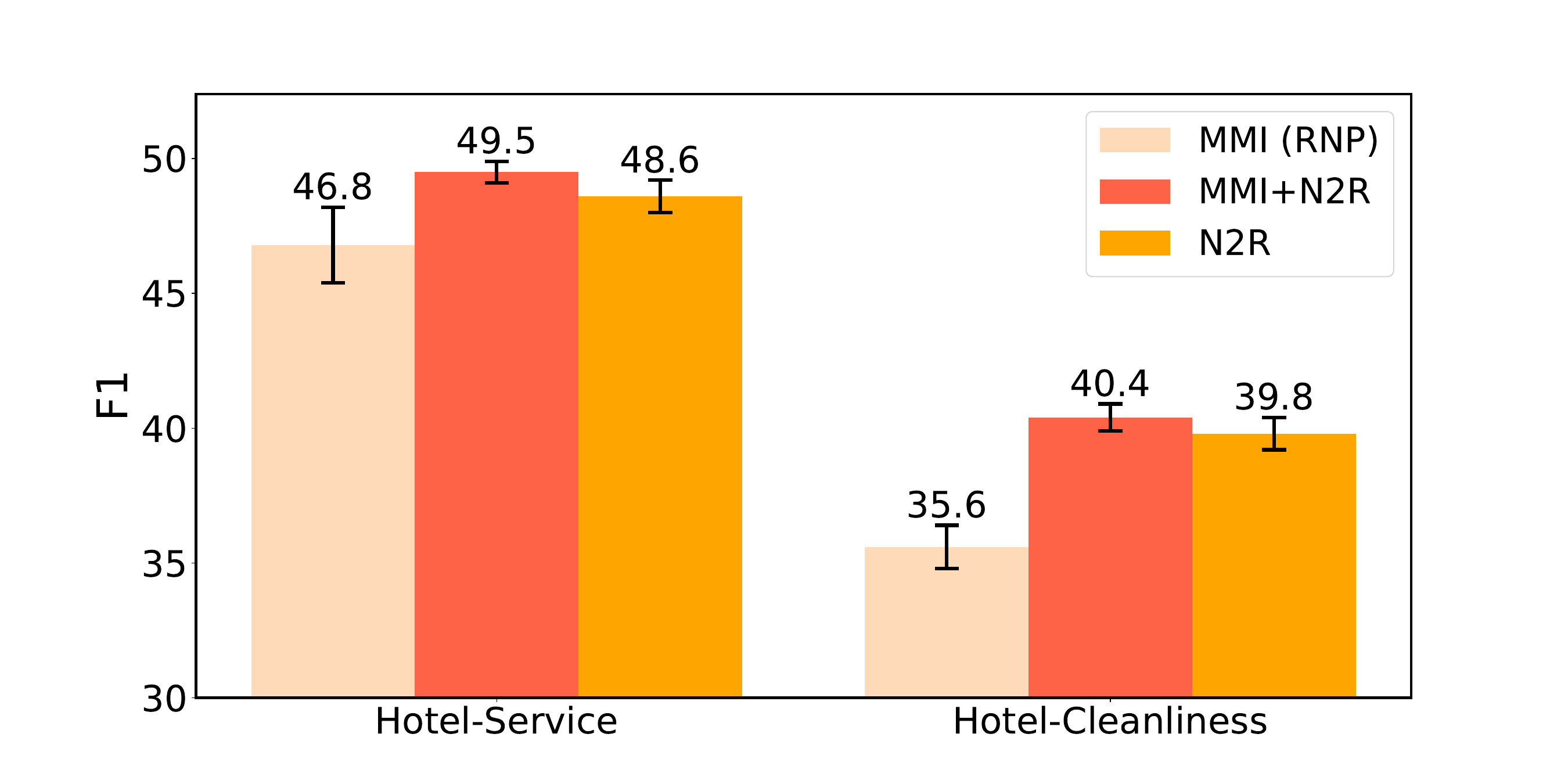}
        \label{fig: hotel noacc}

        }

  \caption{ The comparison between vanilla MMI, N2R, and N2R+MMI on the datasets from (a) BeerAdvocate benchmark and (b) HotelReviews benchmark.
}
% \vspace{-10pt}
  \label{fig: noacc}

\end{figure}

\textbf{N2R can be further improved when combined with MMI}. To further verify the scalability of our N2R, we implement a variant of it by combining N2R and MMI criterion together, which is introduced at the end of $\S$\ref{sec: limitations of mmi}. The comparison between vanilla MMI, N2R, and N2R+MMI on the datasets from the BeerAdvocate benchmark and HotelReviews benchmark are shown in Figure \ref{fig: beer noacc} and Figure \ref{fig: hotel noacc}, respectively.

We can observe that, N2R on its own already significantly outperforms MMI, yet when combined with MMI, its performance improves even further. As compared to the vanilla MMI, the the relevant improvements of MMI+N2R on these four datasets are $13.1\% (=\frac{75.2-66.5}{66.5})$, $8.4\% (=\frac{71.3-65.8}{65.8})$, $5.8\% (=\frac{49.5-46.8}{46.8})$, and $13.5\% (=\frac{40.4-35.6}{35.6})$, respectively. This improvement of MMI+N2R aligns with the empirical results shown in Figure \ref{fig:rationale_acc}. In the initial stage of training, the extractor has not yet identified enough true rationale components, leading to small gradients provided by the OOD objective (Figure \ref{fig:rationale_acc}(c)), which is insufficient to effectively guide the extractor in finding the rationale. However, at this stage, the MMI objective provides larger gradients (Figure \ref{fig:rationale_acc}(b)). In the later stages of training, the situation is reversed. Overall, the MMI and OOD objectives complement each other, resulting in enhanced performance.

\begin{table}

    \caption{Results with GCN encoder on BA2Motifs. }
    \centering
    \vskip 0.1in
    \resizebox{0.6\columnwidth}{!}{
    \setlength\tabcolsep{3pt}
    \begin{tabular}{c|c c | c c | c }
    \hline
        Methods & S &Acc&P&R&F1 \\
\hline
 RNP&20.3 (2.5) &95.2 (1.9)&36.5 (5.5)&36.5 (2.2)& 36.4 (3.8)    \\
 N2R&20.1 (1.2)&96.0 (1.9) & \textbf{40.1} (2.1) & \textbf{40.4} (4.4) & \textbf{40.2} (3.2)\\
 \hline
    \end{tabular}
    }
    \label{tab: ba2motif}
\end{table}

\section{Conclusion and Future Work}
In this paper, we first analyze the diminishing marginal return limitation of the fundamental MMI-based objective in the XAI literature. Then, we propose to use the norm of the intermediate representation of rationale candidates to replace the MMI objective, which is inspired by OOD detection techniques. Our OOD-inspired objective not only outperforms the vanilla MMI, but also beats several recent variants. What's more, it can be easily combined with MMI, further validating its scalability and potential. Our work represents a pioneering attempt to bridge the fields of OOD detection and interpretability. This could potentially inspire researchers in the OOD field to adapt more OOD detection techniques to the XAI domain, thereby further advancing the development of XAI.

Most existing methods find explanations by reconstructing the model’s final output, whereas we focus on identifying which parts of the input are utilized by the network during forward propagation. This is a new direction that can free explanation algorithms from relying on the model's final output. This property could have broader implications, such as potentially being used in the future to explain task-agnostic (i.e., not fine-tuned) pretrained encoders.

\section{Acknowledgment}
This work is supported by the National Key Research and Development Program of China under grant 2024YFC3307900; the National Natural Science Foundation of China under grants 62376103, 62302184, 62436003 and 62206102; Major Science and Technology Project of Hubei Province under grant 2024BAA008; Hubei Science and Technology Talent Service Project under grant 2024DJC078; and Ant Group through CCF-Ant Research Fund. The computation is completed in the HPC Platform of Huazhong University of Science and Technology

We sincerely thank the reviewers and the editor for their valuable feedback and efforts during the review process. They have helped a lot in improving the quality of this paper. We thank Lao Gao and Yuankai Zhang for providing the results on LLMs and image data.

\bibliography{iclr2025_conference}

\begin{thebibliography}{64}
\providecommand{\natexlab}[1]{#1}
\providecommand{\url}[1]{\texttt{#1}}
\expandafter\ifx\csname urlstyle\endcsname\relax
  \providecommand{\doi}[1]{doi: #1}\else
  \providecommand{\doi}{doi: \begingroup \urlstyle{rm}\Url}\fi

\bibitem[Aghajanyan et~al.(2021)Aghajanyan, Gupta, and Zettlemoyer]{AghajanyanGZ20}
Armen Aghajanyan, Sonal Gupta, and Luke Zettlemoyer.
\newblock Intrinsic dimensionality explains the effectiveness of language model fine-tuning.
\newblock In Chengqing Zong, Fei Xia, Wenjie Li, and Roberto Navigli (eds.), \emph{Proceedings of the 59th Annual Meeting of the Association for Computational Linguistics and the 11th International Joint Conference on Natural Language Processing, {ACL/IJCNLP} 2021, (Volume 1: Long Papers), Virtual Event, August 1-6, 2021}, pp.\  7319--7328. Association for Computational Linguistics, 2021.
\newblock \doi{10.18653/V1/2021.ACL-LONG.568}.
\newblock URL \url{https://doi.org/10.18653/v1/2021.acl-long.568}.

\bibitem[Ainsworth et~al.(2023)Ainsworth, Hayase, and Srinivasa]{gitrebis}
Samuel~K. Ainsworth, Jonathan Hayase, and Siddhartha~S. Srinivasa.
\newblock Git re-basin: Merging models modulo permutation symmetries.
\newblock In \emph{The Eleventh International Conference on Learning Representations, {ICLR} 2023, Kigali, Rwanda, May 1-5, 2023}. OpenReview.net, 2023.
\newblock URL \url{https://openreview.net/forum?id=CQsmMYmlP5T}.

\bibitem[Antognini et~al.(2021)Antognini, Musat, and Faltings]{aaai-multiaspect}
Diego Antognini, Claudiu Musat, and Boi Faltings.
\newblock Multi-dimensional explanation of target variables from documents.
\newblock In \emph{Thirty-Fifth {AAAI} Conference on Artificial Intelligence, {AAAI} 2021, Thirty-Third Conference on Innovative Applications of Artificial Intelligence, {IAAI} 2021, The Eleventh Symposium on Educational Advances in Artificial Intelligence, {EAAI} 2021, Virtual Event, February 2-9, 2021}, pp.\  12507--12515. {AAAI} Press, 2021.
\newblock \doi{10.1609/aaai.v35i14.17483}.
\newblock URL \url{https://doi.org/10.1609/aaai.v35i14.17483}.

\bibitem[Bao et~al.(2018)Bao, Chang, Yu, and Barzilay]{2018rationalegumble}
Yujia Bao, Shiyu Chang, Mo~Yu, and Regina Barzilay.
\newblock Deriving machine attention from human rationales.
\newblock In \emph{Proceedings of the 2018 Conference on Empirical Methods in Natural Language Processing, Brussels, Belgium, October 31 - November 4, 2018}, pp.\  1903--1913. Association for Computational Linguistics, 2018.
\newblock \doi{10.18653/v1/d18-1216}.
\newblock URL \url{https://doi.org/10.18653/v1/d18-1216}.

\bibitem[Bastings et~al.(2019)Bastings, Aziz, and Titov]{hardkuma}
Jasmijn Bastings, Wilker Aziz, and Ivan Titov.
\newblock Interpretable neural predictions with differentiable binary variables.
\newblock In \emph{Proceedings of the 57th Conference of the Association for Computational Linguistics, {ACL} 2019, Florence, Italy, July 28- August 2, 2019, Volume 1: Long Papers}, pp.\  2963--2977. Association for Computational Linguistics, 2019.
\newblock \doi{10.18653/v1/p19-1284}.
\newblock URL \url{https://doi.org/10.18653/v1/p19-1284}.

\bibitem[Bolukbasi et~al.(2021)Bolukbasi, Pearce, Yuan, Coenen, Reif, Vi{\'{e}}gas, and Wattenberg]{emb2}
Tolga Bolukbasi, Adam Pearce, Ann Yuan, Andy Coenen, Emily Reif, Fernanda~B. Vi{\'{e}}gas, and Martin Wattenberg.
\newblock An interpretability illusion for {BERT}.
\newblock \emph{CoRR}, abs/2104.07143, 2021.
\newblock URL \url{https://arxiv.org/abs/2104.07143}.

\bibitem[Cai et~al.(2013)Cai, Fan, and Jiang]{highdimvec}
T~Tony Cai, Jianqing Fan, and Tiefeng Jiang.
\newblock Distributions of angles in random packing on spheres.
\newblock \emph{Journal of Machine Learning Research}, 14\penalty0 (136):\penalty0 1837--1864, 2013.

\bibitem[Chan et~al.(2022)Chan, Sanjabi, Mathias, Tan, Nie, Peng, Ren, and Firooz]{Unirex}
Aaron Chan, Maziar Sanjabi, Lambert Mathias, Liang Tan, Shaoliang Nie, Xiaochang Peng, Xiang Ren, and Hamed Firooz.
\newblock {UNIREX:} {A} unified learning framework for language model rationale extraction.
\newblock In \emph{International Conference on Machine Learning, {ICML} 2022, 17-23 July 2022, Baltimore, Maryland, {USA}}, volume 162 of \emph{Proceedings of Machine Learning Research}, pp.\  2867--2889. {PMLR}, 2022.
\newblock URL \url{https://proceedings.mlr.press/v162/chan22a.html}.

\bibitem[Chang et~al.(2020)Chang, Zhang, Yu, and Jaakkola]{invarant}
Shiyu Chang, Yang Zhang, Mo~Yu, and Tommi~S. Jaakkola.
\newblock Invariant rationalization.
\newblock In \emph{Proceedings of the 37th International Conference on Machine Learning, {ICML} 2020, 13-18 July 2020, Virtual Event}, volume 119 of \emph{Proceedings of Machine Learning Research}, pp.\  1448--1458. {PMLR}, 2020.
\newblock URL \url{http://proceedings.mlr.press/v119/chang20c.html}.

\bibitem[Chen et~al.(2022)Chen, He, Narasimhan, and Chen]{danqi}
Howard Chen, Jacqueline He, Karthik Narasimhan, and Danqi Chen.
\newblock Can rationalization improve robustness?
\newblock In \emph{Proceedings of the 2022 Conference of the North American Chapter of the Association for Computational Linguistics: Human Language Technologies, {NAACL} 2022, Seattle, WA, United States, July 10-15, 2022}, pp.\  3792--3805. Association for Computational Linguistics, 2022.
\newblock \doi{10.18653/v1/2022.naacl-main.278}.
\newblock URL \url{https://doi.org/10.18653/v1/2022.naacl-main.278}.

\bibitem[DeYoung et~al.(2020)DeYoung, Jain, Rajani, Lehman, Xiong, Socher, and Wallace]{eraser}
Jay DeYoung, Sarthak Jain, Nazneen~Fatema Rajani, Eric Lehman, Caiming Xiong, Richard Socher, and Byron~C. Wallace.
\newblock {ERASER}: {A} benchmark to evaluate rationalized {NLP} models.
\newblock In Dan Jurafsky, Joyce Chai, Natalie Schluter, and Joel Tetreault (eds.), \emph{Proceedings of the 58th Annual Meeting of the Association for Computational Linguistics}, pp.\  4443--4458, Online, July 2020. Association for Computational Linguistics.
\newblock \doi{10.18653/v1/2020.acl-main.408}.
\newblock URL \url{https://aclanthology.org/2020.acl-main.408}.

\bibitem[Fernandes et~al.(2022)Fernandes, Treviso, Pruthi, Martins, and Neubig]{scott}
Patrick Fernandes, Marcos Treviso, Danish Pruthi, Andr{\'e} Martins, and Graham Neubig.
\newblock Learning to scaffold: Optimizing model explanations for teaching.
\newblock \emph{Advances in Neural Information Processing Systems}, 35:\penalty0 36108--36122, 2022.

\bibitem[Ghassemi et~al.(2021)Ghassemi, Oakden-Rayner, and Beam]{falsehope}
Marzyeh Ghassemi, Luke Oakden-Rayner, and Andrew~L Beam.
\newblock The false hope of current approaches to explainable artificial intelligence in health care.
\newblock \emph{The Lancet Digital Health}, 3\penalty0 (11):\penalty0 e745--e750, 2021.

\bibitem[Guan et~al.(2022)Guan, Li, Leng, Lin, and Guo]{skimformer}
Yue Guan, Zhengyi Li, Jingwen Leng, Zhouhan Lin, and Minyi Guo.
\newblock Transkimmer: Transformer learns to layer-wise skim.
\newblock In Smaranda Muresan, Preslav Nakov, and Aline Villavicencio (eds.), \emph{Proceedings of the 60th Annual Meeting of the Association for Computational Linguistics (Volume 1: Long Papers), {ACL} 2022, Dublin, Ireland, May 22-27, 2022}, pp.\  7275--7286. Association for Computational Linguistics, 2022.
\newblock \doi{10.18653/V1/2022.ACL-LONG.502}.
\newblock URL \url{https://doi.org/10.18653/v1/2022.acl-long.502}.

\bibitem[Gui et~al.(2023)Gui, Liu, Li, Luo, and Ji]{gui2023joint}
Shurui Gui, Meng Liu, Xiner Li, Youzhi Luo, and Shuiwang Ji.
\newblock Joint learning of label and environment causal independence for graph out-of-distribution generalization.
\newblock \emph{arXiv preprint arXiv:2306.01103}, 2023.

\bibitem[Hase et~al.(2020)Hase, Zhang, Xie, and Bansal]{leakage}
Peter Hase, Shiyue Zhang, Harry Xie, and Mohit Bansal.
\newblock Leakage-adjusted simulatability: Can models generate non-trivial explanations of their behavior in natural language?
\newblock In \emph{Findings of the Association for Computational Linguistics: EMNLP 2020}, pp.\  4351--4367, 2020.

\bibitem[Havrylov et~al.(2019)Havrylov, Kruszewski, and Joulin]{cooperative}
Serhii Havrylov, Germ{\'{a}}n Kruszewski, and Armand Joulin.
\newblock Cooperative learning of disjoint syntax and semantics.
\newblock In \emph{Proceedings of the 2019 Conference of the North American Chapter of the Association for Computational Linguistics: Human Language Technologies, {NAACL-HLT} 2019, Minneapolis, MN, USA, June 2-7, 2019, Volume 1 (Long and Short Papers)}, pp.\  1118--1128. Association for Computational Linguistics, 2019.
\newblock \doi{10.18653/v1/n19-1115}.
\newblock URL \url{https://doi.org/10.18653/v1/n19-1115}.

\bibitem[Huang et~al.(2021)Huang, Chen, Du, and Yang]{dmr}
Yongfeng Huang, Yujun Chen, Yulun Du, and Zhilin Yang.
\newblock Distribution matching for rationalization.
\newblock In \emph{Thirty-Fifth {AAAI} Conference on Artificial Intelligence, {AAAI} 2021, Thirty-Third Conference on Innovative Applications of Artificial Intelligence, {IAAI} 2021, The Eleventh Symposium on Educational Advances in Artificial Intelligence, {EAAI} 2021, Virtual Event, February 2-9, 2021}, pp.\  13090--13097. {AAAI} Press, 2021.
\newblock URL \url{https://ojs.aaai.org/index.php/AAAI/article/view/17547}.

\bibitem[Jain et~al.(2020)Jain, Wiegreffe, Pinter, and Wallace]{jain2020faith}
Sarthak Jain, Sarah Wiegreffe, Yuval Pinter, and Byron~C. Wallace.
\newblock Learning to faithfully rationalize by construction.
\newblock In \emph{Proceedings of the 58th Annual Meeting of the Association for Computational Linguistics, {ACL} 2020, Online, July 5-10, 2020}, pp.\  4459--4473. Association for Computational Linguistics, 2020.
\newblock \doi{10.18653/v1/2020.acl-main.409}.
\newblock URL \url{https://doi.org/10.18653/v1/2020.acl-main.409}.

\bibitem[Ji et~al.(2023)Ji, Lee, Frieske, Yu, Su, Xu, Ishii, Bang, Madotto, and Fung]{surveyofllm}
Ziwei Ji, Nayeon Lee, Rita Frieske, Tiezheng Yu, Dan Su, Yan Xu, Etsuko Ishii, Ye~Jin Bang, Andrea Madotto, and Pascale Fung.
\newblock Survey of hallucination in natural language generation.
\newblock \emph{ACM Computing Surveys}, 55\penalty0 (12):\penalty0 1--38, 2023.

\bibitem[Kang et~al.(2024)Kang, Setlur, Tomlin, and Levine]{lownorm}
Katie Kang, Amrith Setlur, Claire Tomlin, and Sergey Levine.
\newblock Deep neural networks tend to extrapolate predictably.
\newblock In \emph{The Twelfth International Conference on Learning Representations}, 2024.

\bibitem[K{\i}c{\i}man et~al.(2023)K{\i}c{\i}man, Ness, Sharma, and Tan]{causalllm}
Emre K{\i}c{\i}man, Robert Ness, Amit Sharma, and Chenhao Tan.
\newblock Causal reasoning and large language models: Opening a new frontier for causality.
\newblock \emph{arXiv preprint arXiv:2305.00050}, 2023.

\bibitem[Kingma \& Ba(2015)Kingma and Ba]{adam}
Diederik~P. Kingma and Jimmy Ba.
\newblock Adam: {A} method for stochastic optimization.
\newblock In \emph{3rd International Conference on Learning Representations, {ICLR} 2015, San Diego, CA, USA, May 7-9, 2015, Conference Track Proceedings}, 2015.
\newblock URL \url{http://arxiv.org/abs/1412.6980}.

\bibitem[Lee et~al.(2024)Lee, Roy, Xu, Raiman, Shoeybi, Catanzaro, and Ping]{emb1}
Chankyu Lee, Rajarshi Roy, Mengyao Xu, Jonathan Raiman, Mohammad Shoeybi, Bryan Catanzaro, and Wei Ping.
\newblock Nv-embed: Improved techniques for training llms as generalist embedding models.
\newblock \emph{CoRR}, abs/2405.17428, 2024.
\newblock \doi{10.48550/ARXIV.2405.17428}.
\newblock URL \url{https://doi.org/10.48550/arXiv.2405.17428}.

\bibitem[Lei et~al.(2016)Lei, Barzilay, and Jaakkola]{emnlp/LeiBJ16}
Tao Lei, Regina Barzilay, and Tommi~S. Jaakkola.
\newblock Rationalizing neural predictions.
\newblock In \emph{Proceedings of the 2016 Conference on Empirical Methods in Natural Language Processing, {EMNLP} 2016, Austin, Texas, USA, November 1-4, 2016}, pp.\  107--117. The Association for Computational Linguistics, 2016.
\newblock \doi{10.18653/v1/d16-1011}.
\newblock URL \url{https://doi.org/10.18653/v1/d16-1011}.

\bibitem[Li et~al.(2023)Li, Fang, Yang, Wang, Ye, Zhao, and Zhang]{evaluatingchatgpt}
Bo~Li, Gexiang Fang, Yang Yang, Quansen Wang, Wei Ye, Wen Zhao, and Shikun Zhang.
\newblock Evaluating chatgpt's information extraction capabilities: An assessment of performance, explainability, calibration, and faithfulness.
\newblock \emph{arXiv preprint arXiv:2304.11633}, 2023.

\bibitem[Li et~al.(2022)Li, Jiang, Li, and Xia]{li2022backdoor}
Yiming Li, Yong Jiang, Zhifeng Li, and Shu-Tao Xia.
\newblock Backdoor learning: A survey.
\newblock \emph{IEEE Transactions on Neural Networks and Learning Systems}, 2022.

\bibitem[Lipton(2018)]{lipton2016mythos}
Zachary~C Lipton.
\newblock The mythos of model interpretability: In machine learning, the concept of interpretability is both important and slippery.
\newblock \emph{Queue}, 16\penalty0 (3):\penalty0 31--57, 2018.

\bibitem[Liu et~al.(2022)Liu, Wang, Wang, Li, Yue, and Zhang]{liufr}
Wei Liu, Haozhao Wang, Jun Wang, Ruixuan Li, Chao Yue, and YuanKai Zhang.
\newblock {FR}: Folded rationalization with a unified encoder.
\newblock In \emph{Advances in Neural Information Processing Systems}, 2022.
\newblock URL \url{https://openreview.net/forum?id=ZPyKSBaKkiO}.

\bibitem[Liu et~al.(2023{\natexlab{a}})Liu, Wang, Wang, Li, Li, Zhang, and Qiu]{liumgr}
Wei Liu, Haozhao Wang, Jun Wang, Ruixuan Li, Xinyang Li, Yuankai Zhang, and Yang Qiu.
\newblock {MGR:} multi-generator based rationalization.
\newblock In Anna Rogers, Jordan~L. Boyd{-}Graber, and Naoaki Okazaki (eds.), \emph{Proceedings of the 61st Annual Meeting of the Association for Computational Linguistics (Volume 1: Long Papers), {ACL} 2023, Toronto, Canada, July 9-14, 2023}, pp.\  12771--12787. Association for Computational Linguistics, 2023{\natexlab{a}}.
\newblock \doi{10.18653/v1/2023.acl-long.715}.
\newblock URL \url{https://doi.org/10.18653/v1/2023.acl-long.715}.

\bibitem[Liu et~al.(2023{\natexlab{b}})Liu, Wang, Wang, Li, Deng, Zhang, and Qiu]{mcd}
Wei Liu, Jun Wang, Haozhao Wang, Ruixuan Li, Zhiying Deng, Yuankai Zhang, and Yang Qiu.
\newblock D-separation for causal self-explanation.
\newblock In \emph{Advances in Neural Information Processing Systems 36: Annual Conference on Neural Information Processing Systems 2023, NeurIPS 2023, New Orleans, LA, USA, December 10 - 16, 2023}, 2023{\natexlab{b}}.
\newblock URL \url{http://papers.nips.cc/paper\_files/paper/2023/hash/87e82678c0d6e5b729398426f82e9af6-Abstract-Conference.html}.

\bibitem[Liu et~al.(2023{\natexlab{c}})Liu, Wang, Wang, Li, Qiu, Zhang, Han, and Zou]{liudr}
Wei Liu, Jun Wang, Haozhao Wang, Ruixuan Li, Yang Qiu, Yuankai Zhang, Jie Han, and Yixiong Zou.
\newblock Decoupled rationalization with asymmetric learning rates: {A} flexible lipschitz restraint.
\newblock In \emph{Proceedings of the 29th {ACM} {SIGKDD} Conference on Knowledge Discovery and Data Mining, {KDD} 2023, Long Beach, CA, USA, August 6-10, 2023}, pp.\  1535--1547. {ACM}, 2023{\natexlab{c}}.
\newblock \doi{10.1145/3580305.3599299}.
\newblock URL \url{https://doi.org/10.1145/3580305.3599299}.

\bibitem[Liu et~al.(2024{\natexlab{a}})Liu, Deng, Niu, Wang, Wang, Zhang, and Li]{mrd}
Wei Liu, Zhiying Deng, Zhongyu Niu, Jun Wang, Haozhao Wang, YuanKai Zhang, and Ruixuan Li.
\newblock Is the mmi criterion necessary for interpretability? degenerating non-causal features to plain noise for self-rationalization.
\newblock In \emph{The Thirty-eighth Annual Conference on Neural Information Processing Systems}, 2024{\natexlab{a}}.

\bibitem[Liu et~al.(2024{\natexlab{b}})Liu, Wang, Wang, Deng, Zhang, Wang, and Li]{dar}
Wei Liu, Haozhao Wang, Jun Wang, Zhiying Deng, Yuankai Zhang, Cheng Wang, and Ruixuan Li.
\newblock Enhancing the rationale-input alignment for self-explaining rationalization.
\newblock In \emph{40th {IEEE} International Conference on Data Engineering, {ICDE} 2024, Utrecht, The Netherlands, May 13-16, 2024}, pp.\  2218--2230. {IEEE}, 2024{\natexlab{b}}.
\newblock \doi{10.1109/ICDE60146.2024.00176}.
\newblock URL \url{https://doi.org/10.1109/ICDE60146.2024.00176}.

\bibitem[Liu et~al.(2024{\natexlab{c}})Liu, Wang, Wang, Li, Deng, and Zhang]{a2i}
Wei Liu, Jun Wang, Haozhao Wang, Ruixuan Li, Zhiying Deng, and YuanKai Zhang.
\newblock Attacking for inspection and instruction: Debiasing self-explaining text classification, 2024{\natexlab{c}}.
\newblock URL \url{https://openreview.net/forum?id=SdoSUDBWJY}.

\bibitem[Luo et~al.(2020)Luo, Cheng, Xu, Yu, Zong, Chen, and Zhang]{pgexplainer}
Dongsheng Luo, Wei Cheng, Dongkuan Xu, Wenchao Yu, Bo~Zong, Haifeng Chen, and Xiang Zhang.
\newblock Parameterized explainer for graph neural network.
\newblock In \emph{Advances in Neural Information Processing Systems 33: Annual Conference on Neural Information Processing Systems 2020, NeurIPS 2020, December 6-12, 2020, virtual}, 2020.
\newblock URL \url{https://proceedings.neurips.cc/paper/2020/hash/e37b08dd3015330dcbb5d6663667b8b8-Abstract.html}.

\bibitem[McAuley et~al.(2012)McAuley, Leskovec, and Jurafsky]{beer}
Julian~J. McAuley, Jure Leskovec, and Dan Jurafsky.
\newblock Learning attitudes and attributes from multi-aspect reviews.
\newblock In \emph{12th {IEEE} International Conference on Data Mining, {ICDM} 2012, Brussels, Belgium, December 10-13, 2012}, pp.\  1020--1025. {IEEE} Computer Society, 2012.
\newblock \doi{10.1109/ICDM.2012.110}.
\newblock URL \url{https://doi.org/10.1109/ICDM.2012.110}.

\bibitem[Pang \& Lee(2004)Pang and Lee]{movie}
Bo~Pang and Lillian Lee.
\newblock A sentimental education: Sentiment analysis using subjectivity summarization based on minimum cuts.
\newblock In \emph{Proceedings of the 42nd Annual Meeting of the Association for Computational Linguistics (ACL-04)}, pp.\  271--278, 2004.

\bibitem[Paranjape et~al.(2020)Paranjape, Joshi, Thickstun, Hajishirzi, and Zettlemoyer]{informationbottle}
Bhargavi Paranjape, Mandar Joshi, John Thickstun, Hannaneh Hajishirzi, and Luke Zettlemoyer.
\newblock An information bottleneck approach for controlling conciseness in rationale extraction.
\newblock In \emph{Proceedings of the 2020 Conference on Empirical Methods in Natural Language Processing, {EMNLP} 2020, Online, November 16-20, 2020}, pp.\  1938--1952. Association for Computational Linguistics, 2020.
\newblock \doi{10.18653/v1/2020.emnlp-main.153}.
\newblock URL \url{https://doi.org/10.18653/v1/2020.emnlp-main.153}.

\bibitem[Plyler et~al.(2021)Plyler, Green, and Chi]{counter}
Mitchell Plyler, Michael Green, and Min Chi.
\newblock Making a (counterfactual) difference one rationale at a time.
\newblock In \emph{Advances in Neural Information Processing Systems 34: Annual Conference on Neural Information Processing Systems 2021, NeurIPS 2021, December 6-14, 2021, virtual}, pp.\  28701--28713, 2021.
\newblock URL \url{https://proceedings.neurips.cc/paper/2021/hash/f0f800c92d191d736c4411f3b3f8ef4a-Abstract.html}.

\bibitem[Pradhan et~al.(2022)Pradhan, Zhu, Glavic, and Salimi]{sigmod-debug}
Romila Pradhan, Jiongli Zhu, Boris Glavic, and Babak Salimi.
\newblock Interpretable data-based explanations for fairness debugging.
\newblock In \emph{{SIGMOD} '22: International Conference on Management of Data, Philadelphia, PA, USA, June 12 - 17, 2022}, pp.\  247--261. {ACM}, 2022.
\newblock \doi{10.1145/3514221.3517886}.
\newblock URL \url{https://doi.org/10.1145/3514221.3517886}.

\bibitem[Qin et~al.(2023)Qin, Zhang, Zhang, Chen, Yasunaga, and Yang]{chatgptgeneral}
Chengwei Qin, Aston Zhang, Zhuosheng Zhang, Jiaao Chen, Michihiro Yasunaga, and Diyi Yang.
\newblock Is chatgpt a general-purpose natural language processing task solver?
\newblock \emph{arXiv preprint arXiv:2302.06476}, 2023.

\bibitem[Rajagopal et~al.(2021)Rajagopal, Balachandran, Hovy, and Tsvetkov]{concept}
Dheeraj Rajagopal, Vidhisha Balachandran, Eduard~H Hovy, and Yulia Tsvetkov.
\newblock {SELFEXPLAIN}: A self-explaining architecture for neural text classifiers.
\newblock In \emph{Proceedings of the 2021 Conference on Empirical Methods in Natural Language Processing}, pp.\  836--850, Online and Punta Cana, Dominican Republic, November 2021. Association for Computational Linguistics.
\newblock \doi{10.18653/v1/2021.emnlp-main.64}.
\newblock URL \url{https://aclanthology.org/2021.emnlp-main.64}.

\bibitem[Ren et~al.(2024)Ren, Gao, Shen, and Zhang]{ren2024}
Qihan Ren, Jiayang Gao, Wen Shen, and Quanshi Zhang.
\newblock Where we have arrived in proving the emergence of sparse interaction primitives in {DNN}s.
\newblock In \emph{The Twelfth International Conference on Learning Representations}, 2024.
\newblock URL \url{https://openreview.net/forum?id=3pWSL8My6B}.

\bibitem[Rudin(2019)]{stopposthoc}
Cynthia Rudin.
\newblock Stop explaining black box machine learning models for high stakes decisions and use interpretable models instead.
\newblock \emph{Nature Machine Intelligence}, 1\penalty0 (5):\penalty0 206--215, 2019.
\newblock \doi{10.1038/S42256-019-0048-X}.
\newblock URL \url{https://doi.org/10.1038/s42256-019-0048-x}.

\bibitem[Sagawa et~al.(2019)Sagawa, Koh, Hashimoto, and Liang]{waterbird}
Shiori Sagawa, Pang~Wei Koh, Tatsunori~B Hashimoto, and Percy Liang.
\newblock Distributionally robust neural networks for group shifts: On the importance of regularization for worst-case generalization.
\newblock \emph{arXiv preprint arXiv:1911.08731}, 2019.

\bibitem[Seiler(2023)]{datacentric}
Benjamin~B Seiler.
\newblock \emph{Applications of Cooperative Game Theory to Interpretable Machine Learning}.
\newblock PhD thesis, Stanford University, 2023.

\bibitem[Sha et~al.(2021)Sha, Camburu, and Lukasiewicz]{AAAI21learningfrombest}
Lei Sha, Oana{-}Maria Camburu, and Thomas Lukasiewicz.
\newblock Learning from the best: Rationalizing predictions by adversarial information calibration.
\newblock In \emph{Thirty-Fifth {AAAI} Conference on Artificial Intelligence, {AAAI} 2021, Thirty-Third Conference on Innovative Applications of Artificial Intelligence, {IAAI} 2021, The Eleventh Symposium on Educational Advances in Artificial Intelligence, {EAAI} 2021, Virtual Event, February 2-9, 2021}, pp.\  13771--13779. {AAAI} Press, 2021.
\newblock URL \url{https://ojs.aaai.org/index.php/AAAI/article/view/17623}.

\bibitem[Storek et~al.(2023)Storek, Subbiah, and McKeown]{noise}
Adam Storek, Melanie Subbiah, and Kathleen~R. McKeown.
\newblock Unsupervised selective rationalization with noise injection.
\newblock In \emph{Proceedings of the 61st Annual Meeting of the Association for Computational Linguistics (Volume 1: Long Papers), {ACL} 2023, Toronto, Canada, July 9-14, 2023}, pp.\  12647--12659. Association for Computational Linguistics, 2023.
\newblock \doi{10.18653/v1/2023.acl-long.707}.
\newblock URL \url{https://doi.org/10.18653/v1/2023.acl-long.707}.

\bibitem[Sun et~al.(2024)Sun, Huang, Wang, Wu, Zhang, Gao, Huang, Lyu, Zhang, Li, Liu, Liu, Wang, Zhang, Kailkhura, Xiong, Zhang, Xiao, Li, Xing, Huang, Liu, Ji, Wang, Zhang, Yao, Kellis, Zitnik, Jiang, Bansal, Zou, Pei, Liu, Gao, Han, Zhao, Tang, Wang, Mitchell, Shu, Xu, Chang, He, Huang, Backes, Gong, Yu, Chen, Gu, Xu, Ying, Ji, Jana, Chen, Liu, Zhou, Wang, Li, Zhang, Wang, Xie, Chen, Wang, Liu, Ye, Cao, and Zhao]{trustllm}
Lichao Sun, Yue Huang, Haoran Wang, Siyuan Wu, Qihui Zhang, Chujie Gao, Yixin Huang, Wenhan Lyu, Yixuan Zhang, Xiner Li, Zhengliang Liu, Yixin Liu, Yijue Wang, Zhikun Zhang, Bhavya Kailkhura, Caiming Xiong, Chao Zhang, Chaowei Xiao, Chunyuan Li, Eric~P. Xing, Furong Huang, Hao Liu, Heng Ji, Hongyi Wang, Huan Zhang, Huaxiu Yao, Manolis Kellis, Marinka Zitnik, Meng Jiang, Mohit Bansal, James Zou, Jian Pei, Jian Liu, Jianfeng Gao, Jiawei Han, Jieyu Zhao, Jiliang Tang, Jindong Wang, John Mitchell, Kai Shu, Kaidi Xu, Kai{-}Wei Chang, Lifang He, Lifu Huang, Michael Backes, Neil~Zhenqiang Gong, Philip~S. Yu, Pin{-}Yu Chen, Quanquan Gu, Ran Xu, Rex Ying, Shuiwang Ji, Suman Jana, Tianlong Chen, Tianming Liu, Tianyi Zhou, William Wang, Xiang Li, Xiangliang Zhang, Xiao Wang, Xing Xie, Xun Chen, Xuyu Wang, Yan Liu, Yanfang Ye, Yinzhi Cao, and Yue Zhao.
\newblock Trustllm: Trustworthiness in large language models.
\newblock \emph{CoRR}, abs/2401.05561, 2024.
\newblock \doi{10.48550/ARXIV.2401.05561}.
\newblock URL \url{https://doi.org/10.48550/arXiv.2401.05561}.

\bibitem[Wah et~al.(2011)Wah, Branson, Welinder, Perona, and Belongie]{cub}
Catherine Wah, Steve Branson, Peter Welinder, Pietro Perona, and Serge Belongie.
\newblock The caltech-ucsd birds-200-2011 dataset.
\newblock 2011.

\bibitem[Wang et~al.(2010)Wang, Lu, and Zhai]{hotel}
Hongning Wang, Yue Lu, and Chengxiang Zhai.
\newblock Latent aspect rating analysis on review text data: a rating regression approach.
\newblock In \emph{Proceedings of the 16th {ACM} {SIGKDD} International Conference on Knowledge Discovery and Data Mining, Washington, DC, USA, July 25-28, 2010}, pp.\  783--792. {ACM}, 2010.
\newblock \doi{10.1145/1835804.1835903}.
\newblock URL \url{https://doi.org/10.1145/1835804.1835903}.

\bibitem[Wei et~al.(2022)Wei, Wang, Schuurmans, Bosma, Ichter, Xia, Chi, Le, and Zhou]{cot}
Jason Wei, Xuezhi Wang, Dale Schuurmans, Maarten Bosma, Brian Ichter, Fei Xia, Ed~H. Chi, Quoc~V. Le, and Denny Zhou.
\newblock Chain-of-thought prompting elicits reasoning in large language models.
\newblock In \emph{NeurIPS}, 2022.
\newblock URL \url{http://papers.nips.cc/paper\_files/paper/2022/hash/9d5609613524ecf4f15af0f7b31abca4-Abstract-Conference.html}.

\bibitem[Wu et~al.(2022)Wu, Wang, Zhang, He, and Chua]{dir}
Yingxin Wu, Xiang Wang, An~Zhang, Xiangnan He, and Tat{-}Seng Chua.
\newblock Discovering invariant rationales for graph neural networks.
\newblock In \emph{The Tenth International Conference on Learning Representations, {ICLR} 2022, Virtual Event, April 25-29, 2022}. OpenReview.net, 2022.
\newblock URL \url{https://openreview.net/forum?id=hGXij5rfiHw}.

\bibitem[Xia et~al.(2024)Xia, Malladi, Gururangan, Arora, and Chen]{less}
Mengzhou Xia, Sadhika Malladi, Suchin Gururangan, Sanjeev Arora, and Danqi Chen.
\newblock {LESS:} selecting influential data for targeted instruction tuning.
\newblock \emph{CoRR}, abs/2402.04333, 2024.
\newblock \doi{10.48550/ARXIV.2402.04333}.
\newblock URL \url{https://doi.org/10.48550/arXiv.2402.04333}.

\bibitem[Ye et~al.(2023)Ye, Chen, Xu, Zu, Shao, Liu, Cui, Zhou, Gong, Shen, et~al.]{comprehensivechatgpt}
Junjie Ye, Xuanting Chen, Nuo Xu, Can Zu, Zekai Shao, Shichun Liu, Yuhan Cui, Zeyang Zhou, Chao Gong, Yang Shen, et~al.
\newblock A comprehensive capability analysis of gpt-3 and gpt-3.5 series models.
\newblock \emph{arXiv preprint arXiv:2303.10420}, 2023.

\bibitem[Ying et~al.(2019)Ying, Bourgeois, You, Zitnik, and Leskovec]{gnnexplainer}
Zhitao Ying, Dylan Bourgeois, Jiaxuan You, Marinka Zitnik, and Jure Leskovec.
\newblock Gnnexplainer: Generating explanations for graph neural networks.
\newblock In \emph{Advances in Neural Information Processing Systems 32: Annual Conference on Neural Information Processing Systems 2019, NeurIPS 2019, December 8-14, 2019, Vancouver, BC, Canada}, pp.\  9240--9251, 2019.
\newblock URL \url{https://proceedings.neurips.cc/paper/2019/hash/d80b7040b773199015de6d3b4293c8ff-Abstract.html}.

\bibitem[Yu et~al.(2019)Yu, Chang, Zhang, and Jaakkola]{rethinking}
Mo~Yu, Shiyu Chang, Yang Zhang, and Tommi~S. Jaakkola.
\newblock Rethinking cooperative rationalization: Introspective extraction and complement control.
\newblock In \emph{Proceedings of the 2019 Conference on Empirical Methods in Natural Language Processing and the 9th International Joint Conference on Natural Language Processing, {EMNLP-IJCNLP} 2019, Hong Kong, China, November 3-7, 2019}, pp.\  4092--4101. Association for Computational Linguistics, 2019.
\newblock \doi{10.18653/v1/D19-1420}.
\newblock URL \url{https://doi.org/10.18653/v1/D19-1420}.

\bibitem[Yu et~al.(2021)Yu, Zhang, Chang, and Jaakkola]{interlocking}
Mo~Yu, Yang Zhang, Shiyu Chang, and Tommi~S. Jaakkola.
\newblock Understanding interlocking dynamics of cooperative rationalization.
\newblock In \emph{Advances in Neural Information Processing Systems 34: Annual Conference on Neural Information Processing Systems 2021, NeurIPS 2021, December 6-14, 2021, virtual}, pp.\  12822--12835, 2021.
\newblock URL \url{https://proceedings.neurips.cc/paper/2021/hash/6a711a119a8a7a9f877b5f379bfe9ea2-Abstract.html}.

\bibitem[Yuan et~al.(2022)Yuan, Cai, Hu, Wang, and Ji]{GDM}
Hao Yuan, Lei Cai, Xia Hu, Jie Wang, and Shuiwang Ji.
\newblock Interpreting image classifiers by generating discrete masks.
\newblock \emph{{IEEE} Trans. Pattern Anal. Mach. Intell.}, 44\penalty0 (4):\penalty0 2019--2030, 2022.
\newblock \doi{10.1109/TPAMI.2020.3028783}.
\newblock URL \url{https://doi.org/10.1109/TPAMI.2020.3028783}.

\bibitem[Yue et~al.(2023)Yue, Liu, Wang, An, Du, and Huang]{interventional}
Linan Yue, Qi~Liu, Li~Wang, Yanqing An, Yichao Du, and Zhenya Huang.
\newblock Interventional rationalization.
\newblock In \emph{Proceedings of the 2016 Conference on Empirical Methods in Natural Language Processing, {EMNLP} 2023, Singapore, December 6 –10, 2023}, 2023.
\newblock URL \url{https://openreview.net/forum?id=KoEa6h1o6D1}.

\bibitem[Zhang et~al.(2023)Zhang, Wu, Wang, Cai, and Cai]{cr}
Wenbo Zhang, Tong Wu, Yunlong Wang, Yong Cai, and Hengrui Cai.
\newblock Towards trustworthy explanation: On causal rationalization.
\newblock In \emph{Proceedings of the 40th International Conference on Machine Learning (ICML'23)}, volume 202 of \emph{Proceedings of Machine Learning Research}, pp.\  41715--41736. PMLR, 23--29 Jul 2023.
\newblock URL \url{https://arxiv.org/abs/2306.14115}.

\bibitem[Zhang et~al.(2024)Zhang, Kong, Wang, Li, Wang, Li, and Liu]{ecairationale}
Yuankai Zhang, Lingxiao Kong, Haozhao Wang, Ruixuan Li, Jun Wang, Yuhua Li, and Wei Liu.
\newblock Adversarial attack for explanation robustness of rationalization models.
\newblock In \emph{{ECAI} 2024 - 27th European Conference on Artificial Intelligence, 19-24 October 2024, Santiago de Compostela, Spain - Including 13th Conference on Prestigious Applications of Intelligent Systems {(PAIS} 2024)}, volume 392 of \emph{Frontiers in Artificial Intelligence and Applications}, pp.\  2290--2297. {IOS} Press, 2024.
\newblock \doi{10.3233/FAIA240752}.
\newblock URL \url{https://doi.org/10.3233/FAIA240752}.

\bibitem[Zhao et~al.(2024)Zhao, Wang, Li, Liang, and Li]{agr}
Yunxiao Zhao, Zhiqiang Wang, Xiaoli Li, Jiye Liang, and Ru~Li.
\newblock {AGR:} reinforced causal agent-guided self-explaining rationalization.
\newblock In \emph{Proceedings of the 62nd Annual Meeting of the Association for Computational Linguistics, {ACL} 2024 - Short Papers, Bangkok, Thailand, August 11-16, 2024}, pp.\  510--518. Association for Computational Linguistics, 2024.
\newblock URL \url{https://aclanthology.org/2024.acl-short.47}.

\end{thebibliography}
\bibliographystyle{iclr2025_conference}

\clearpage
\appendix
\section{More Details}
\subsection{Datasets}\label{app: datasets} 
The statistics of the datasets are in Table \ref{tab:dataset}. $Pos$ and $Neg$ denote the number of positive and negative examples in each set. $S$ denotes the average percentage of tokens in human-annotated rationales to the whole texts.
\begin{table}[t]
\caption{Statistics of datasets used in this paper}
   \centering
       \vskip 0.1in
   % \footnotesize
   % \setlength\tabcolsep{2pt}
    \begin{tabular}{c l| c c| c c| c c c}
    \hline
         \multicolumn{2}{c|}{\multirow{2}{*}{Datasets}}&\multicolumn{2}{c|}{Train}&\multicolumn{2}{c|}{Dev}&\multicolumn{3}{c}{Annotation}  \\
         \multicolumn{2}{c|}{}& Pos&Neg&Pos&Neg&Pos&Neg&S\\
         \hline\hline
        \multirow{2}{*}{Beer}&Appearance&16891&16891 &6628&2103&923&13&18.5\\
        {}&Aroma&15169&15169&6579&2218&848&29&15.6\\
        \hline\hline
        \multirow{2}{*}{Hotel}
        {}&Service&50742&50742&6344&6344&101&99&11.5\\
        {}&Cleanliness&75049&75049&9382&9382&99&101&8.9\\
        \hline
    \end{tabular}
    
    \label{tab:dataset}
\end{table}

\subsection{Experimental details}\label{app: experimental details}

The code and detailed running instructions will be made publicly available on GitHub after the paper is accepted. The code is now in an anonymous repository: \url{https://anonymous.4open.science/r/N2R-0E5E}. The anonymous repository will be closed in Nov 13, as we do not want people other than the reviewers to get the code.

The maximum sequence length is set to 256. We use the Adam optimizer \cite{adam} with its default parameters, except for the learning rate (the learning rate is 0.0001). The temperature for gumbel-softmax is the default value $1$. 
We implement the code with Pytorch on a RTX4090 GPU. We report the average results of five random seeds, and the seeds are [1,2,3,4,5].

For NIR and our N2R, considering they are both variants of the standard RNP, we first manually tune the hyperparameters for RNP, and then apply the hyperparameters to both NIR and N2R.
For all datasets, we use a learning rate of 0.0001. The batchsize is 128 for the beer-related datasets and 256 for the hotel-related datasets. These hyperparameters are found by manually tune the standard RNP and are applied to both NIR and our N2R.

The core idea of NIR is to inject noise into the selected rationales. We use RNP as its backbone. A unique hyperparameter of NIR is the proportion of noise. Following the method in its original paper, we searched within $[0.1, 0.2, 0.3]$ and found that $0.1$ yielded the best results on most datasets, hence we adopted $0.1$ for it.

We found that the training of Inter\_RAT is very unstable. To avoid potential unfair factors, our main settings are determined with reference to it. Except for the part about sparsity, we used its original hyperparameters for it.

For A2I, we contact its authors and get the code and hyperparameters of it.

For CR, we just keep the major settings (``bert-base-uncased", the Beer-Appearance dataset, and the sprasity of $10\%$) the same as it and copy its results from its original paper. 

\subsection{Implementation details of N2R} \label{app: implementation of N2R}

For a batch of $(X,Y)$, we first send $X$ to the extractor and get the rationale $Z$:
\begin{equation}
    Z=f_E(X).
\end{equation}
Then, we get a copy of $Z$ with the pytorch function $``$torch.detach()$"$:
\begin{equation}
    Z'=\text{torch.detach}(Z),
\end{equation}
such that the following computation does not involve the extractor's gradients.
Then, we send the rationale $Z'$ to the predictor and get the prediction $\hat{Y}$:
\begin{equation}
    \hat{Y}=f_P(Z').
\end{equation}
Then, we update the predictor with the cross-entropy loss.
\begin{equation}
    \mathop{\min}_{\theta_p}H_c(Y,\hat{Y})
\end{equation}
Note that this updating process with cross-entropy will not influence the extractor, since we have used $``$torch.detach()$"$ for $Z$.

Then, we fix the parameters of the predictor, and only update the extractor. We first get the rationale candidate $Z$ with 
\begin{equation}
    Z=f_E(X).
\end{equation}
And we then send it to the predictor's encoder to get $||Enc(Z)||_2$.
Then, we update the extractor with 
\begin{equation}
   \mathop{\min}_{\theta_e} -log (||Enc(Z)||_2).  
\end{equation}

Then, we get into the next round to update the extractor and the predictor again.

\begin{table*}[t]
    \centering

       \caption{The quality of rationales extracted by llama-3.1-8b-instruct.}
               \vskip 0.1in
       {\resizebox{0.99\columnwidth}{!}{
       
    \begin{tabular}{c c c |c | c c c| c| c c c }
\hline
\multicolumn{3}{c|}{\multirow{2}{*}{\diagbox{Methods}{Datasets}}} & \multicolumn{4}{c|}{Beer-Appearance} & \multicolumn{4}{c}{Beer-Aroma} \\
\cline{4-11}
\multicolumn{3}{c|}{} &S & P & R &\multicolumn{1}{c|}{F1} &S&  P & R &\multicolumn{1}{c}{F1} \\
\hline

\multicolumn{3}{c|}{N2R (ours)} & 14.8 (0.5) & 81.9 (2.7) & 65.3 (2.2) & \textbf{72.7} (2.1) & 14.9 (0.4) & 70.2 (1.5) & 67.2 (1.3) & \textbf{68.7} (1.1)\\

\multicolumn{3}{c|}{llama-3.1-8b (finetune) } & n/a& 86.3 (n/a)&46.2 (n/a)& 60.2(n/a)& n/a & 73.2 (n/a)& 50.6 (n/a) &59.8 (n/a)\\
\multicolumn{3}{c|}{llama-3.1-8b (2 shot) } & n/a& 15.4 (n/a) & 16.0 (n/a)& 15.7 (n/a)& n/a & 17.9 (n/a)& 24.2 (n/a) & 20.6 (n/a)\\

\hline

\\
\\
\hline
\multicolumn{3}{c|}{\multirow{2}{*}{\diagbox{Methods}{Datasets}}} & \multicolumn{4}{c|}{Hotel-Service} & \multicolumn{4}{c}{Hotel-Cleanliness} \\
\cline{4-11}
\multicolumn{3}{c|}{} &S & P & R &\multicolumn{1}{c|}{F1} &S&  P & R &\multicolumn{1}{c}{F1} \\
\hline

\multicolumn{3}{c|}{N2R (ours)} & 15.1 (0.2)& 42.8 (0.5) & 56.3 (1.0) & 48.6 (0.6) & 14.8 (0.2) & 31.8 (0.5) & 53.4 (0.8) & 39.8 (0.6)\\

\multicolumn{3}{c|}{llama-3.1-8b (finetune) } & n/a &77.3 (n/a)&40.6 (n/a)&\textbf{53.3} (n/a)&n/a &54.9 (n/a)&31.3 (n/a)&39.9 (n/a)\\

\multicolumn{3}{c|}{llama-3.1-8b (2 shot) } & n/a & 45.3 (n/a)&51.7 (n/a)&48.3 (n/a)&n/a & 39.3 (n/a)&43.0 (n/a)&\textbf{41.1} (n/a)\\
\hline

\end{tabular}
}

} 
    \label{tab: results of llm}

\end{table*}

\subsection{Implementation details of N2R+MMI}\label{app: implementation of N2R+RNP}
The implementation details are similar to those of Appendix \ref{app: implementation of N2R}. The only difference is that we no more use ``torch.detach''. 

\subsection{The minimum cross-entropy is equal to the entropy}\label{app: minimizing cross entropy is equal to kl}

The cross-entropy consists of two parts:
\begin{equation}
    H_c(Y,\hat{Y}|Z)=H(Y|Z)+D_{KL}(P(Y|Z)||P(\hat{Y}|Z)).
\end{equation}
When we minimizing the cross-entropy $H_c(Y,\hat{Y}|Z)$ by adjusting the predictor's parameters, we are in fact minimizing $D_{KL}(P(Y|Z)||P(\hat{Y}|Z))$.
And we know that if the predictor is trained ideally, we have $\min D_{KL}(P(Y|Z)||P(\hat{Y}|Z))=0$. Then, we have 
\begin{equation}
    \mathop{\min}_{\theta_p}H_c(Y,\hat{Y}|Z)=H(Y|Z).
\end{equation}

\subsection{The detailed discussion about Equation (\ref{eqa: mmi with two variable})}\label{app: proof of mmi with two variable}
In any cases, we have 
\begin{equation}
    I(Y;R_1,R_2)=I(Y;R_1)+I(Y;R_2|R1).
\end{equation}
\begin{equation}
    I(Y;R_1,R_2)-I(Y;R_1)-I(Y;R_2)=I(Y;R_2|R1)-I(Y;R_2).
\end{equation}
The magnitude relationship between $I(Y;R_2|R1)$ and $I(Y;R_2)$ is arbitrary. In other words, there exists some scenarios where $I(Y;R_2|R1)<I(Y;R_2)$. In such cases, we have 
\begin{equation}
    I(Y;R_1,R_2)<I(Y;R_1)+I(Y;R_2).
\end{equation}
In these cases, the mutual information faces the diminishing marginal returns problem.

\subsection{The rationales extracted by llama-3.1-8b-instruct}\label{app: llm result}

\begin{wrapfigure}[16]{R}{0.5\columnwidth}
    \centering
    \includegraphics[width=0.5\columnwidth]{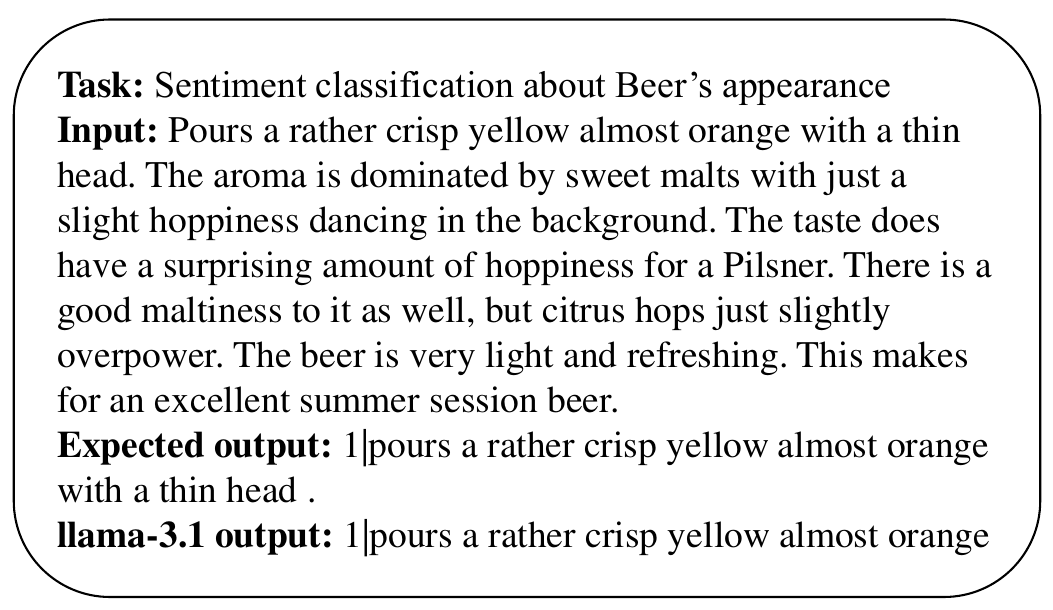}

    \caption{An example of llama's output. Here ``1'' means that the class label $Y$ is positive. And the words after ``$|$'' represent the rationale. }
    \label{fig: example of llm}

\end{wrapfigure}

To further show the potential impact of rationalization in the era of LLMs, here we present the results of the experiments conducted with the llama-3.1-8b-instruct model. We perform both 2-shot prompting and supervised fine-tuning. 

For 2-shot prompting, we provide the model with a negative text with its corresponding rationale, and a positive text with its corresponding rationale. For supervised fine-tuning, the supervison label is the classification label, since we perform unsupervised rationale extraction.  We use 4*RTX 4090 24GB GPUs and LoRA to fine tune the models. We provide a detailed document in our anonymous code repository (\url{https://anonymous.4open.science/r/N2R-0E5E/details_of_llms.pdf}) to include all the details (including the prompt templates, LoRA fine-tuning parameter settings, and more).

    \begin{wrapfigure}[21]{R}{0.5\columnwidth}
    \centering
    \includegraphics[width=0.5\columnwidth]{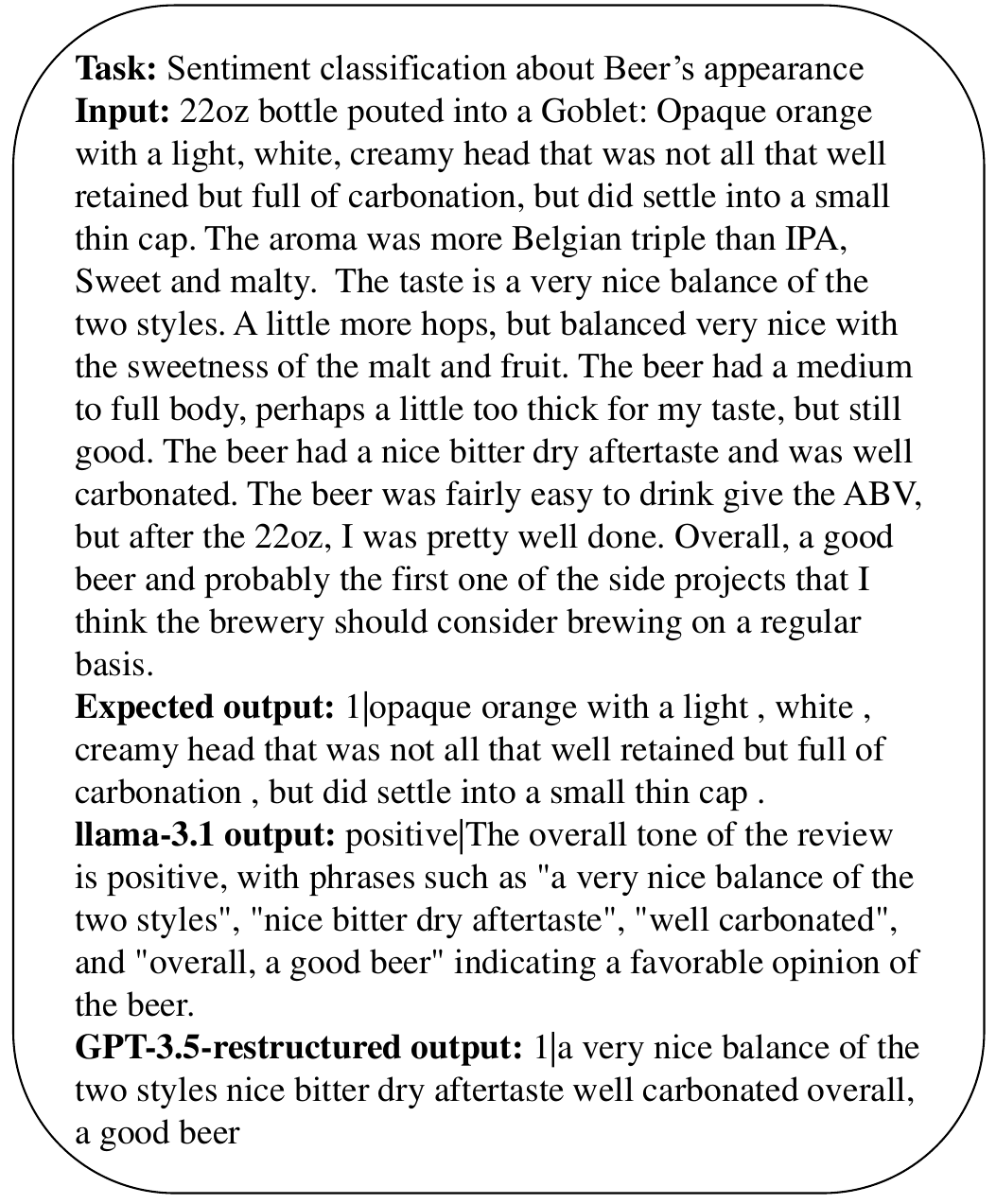}
    \caption{An example of llama fails to output the rationale in the right format.  }
    \label{fig: fail of llm}

\end{wrapfigure}

In most cases, the model can output the rationale in the correct format. Figure \ref{fig: example of llm} shows an example. But in 2-shot prompting, the model sometimes outputs additional parts along with the rationale (through manual observation, this situation does not occur frequently.). Figure \ref{fig: fail of llm} is another example. In such cases, we use gpt-3.5-turbo to extract the content within the quotation marks. 

The results are shown in Table \ref{tab: results of llm}. LLMs are not good at counting, so we did not constrain the percentage length (i.e., sparsity) of the rationale extracted by the model. Compared to our N2R, llama-3.1 does not have a crushing advantage. On two out of four datasets, our N2R outperforms  llama-3.1. And on the left two datasets, N2R achieves comparable  results to llama-3.1. Besides, our N2R can be applied to graph data, while it is not easy to do so for LLMs.

\subsection{More results of diminishing  marginal returns}\label{app: marginal return}
\begin{figure*}[t]
    % \flushleft
      \subfigure[]{
            \includegraphics[width=0.31\columnwidth]{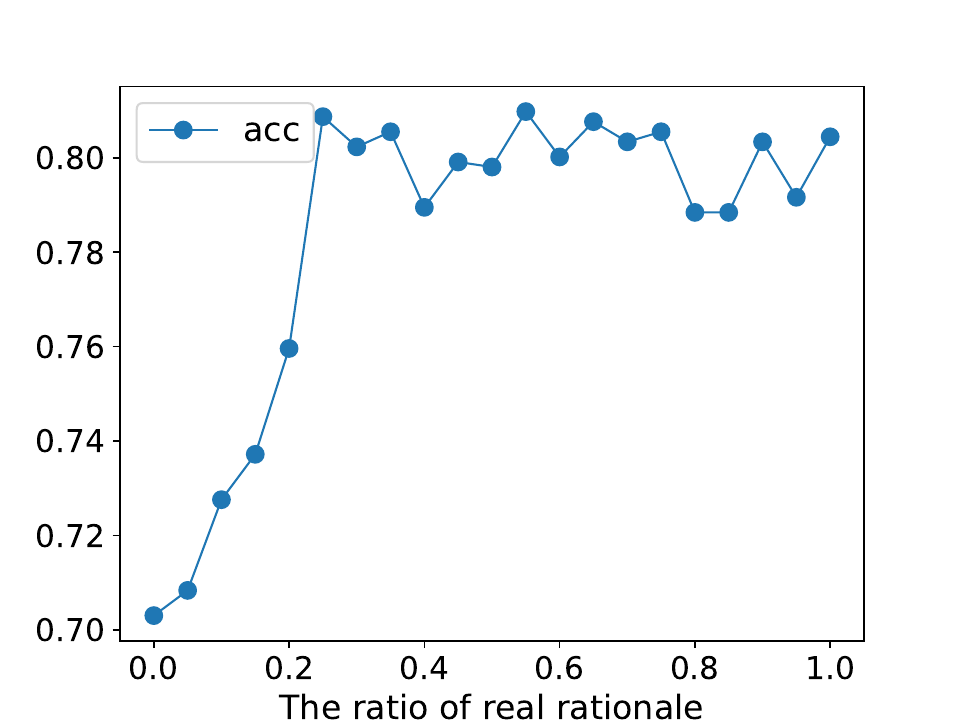}
        \label{fig:rationale_acc1 beer0}
        }
        \subfigure[]{
            \includegraphics[width=0.31\columnwidth]{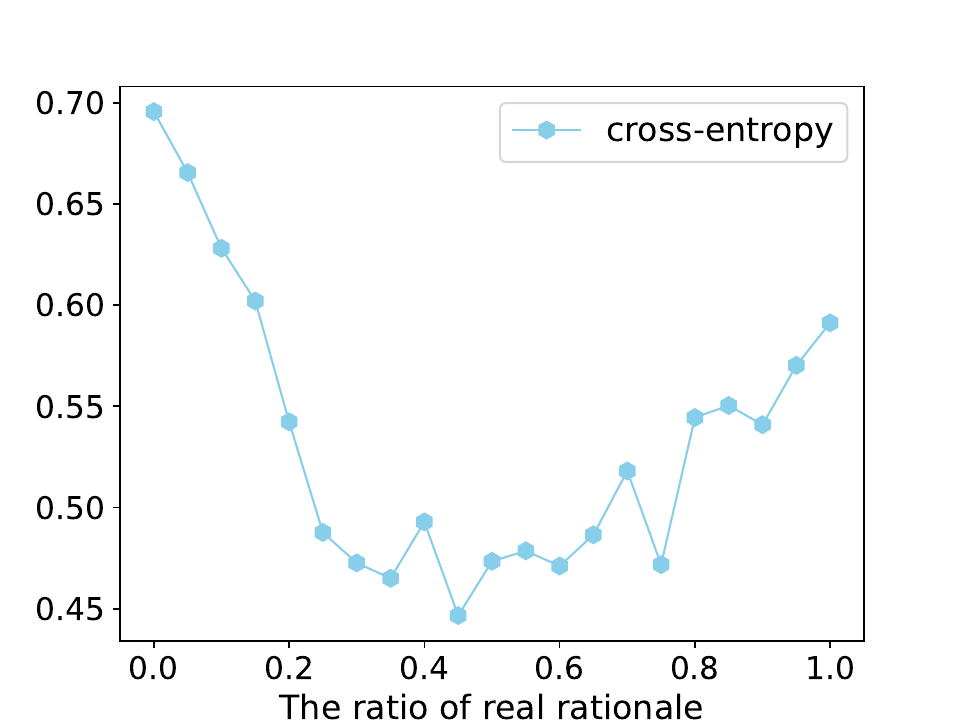}
            \label{fig:rationale_loss beer0}
        }
        \subfigure[]{
            \includegraphics[width=0.31\columnwidth]{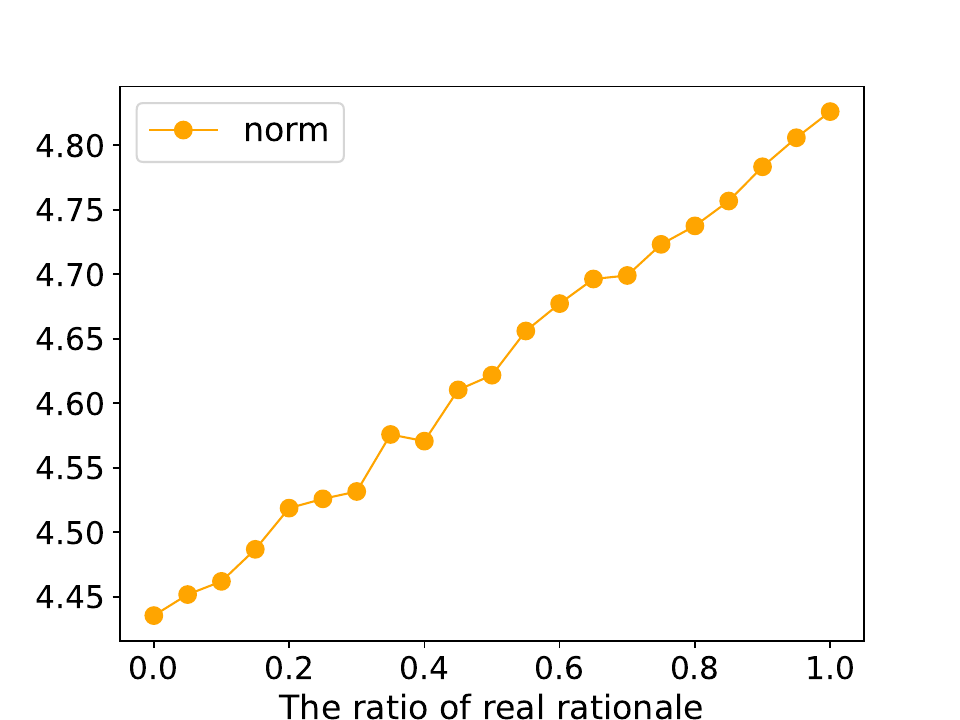}
            \label{fig:rationale_norm beer0}
        }
        \subfigure[]{
            \includegraphics[width=0.31\columnwidth]{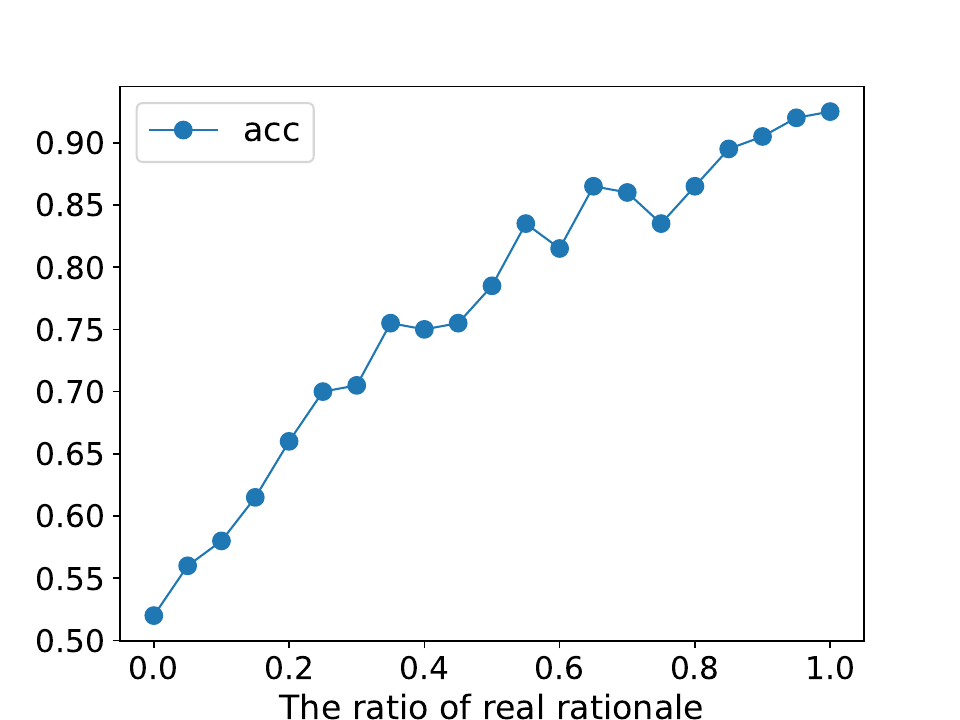}
        \label{fig:rationale_acc1 hotel1}
        }
        \subfigure[]{
            \includegraphics[width=0.31\columnwidth]{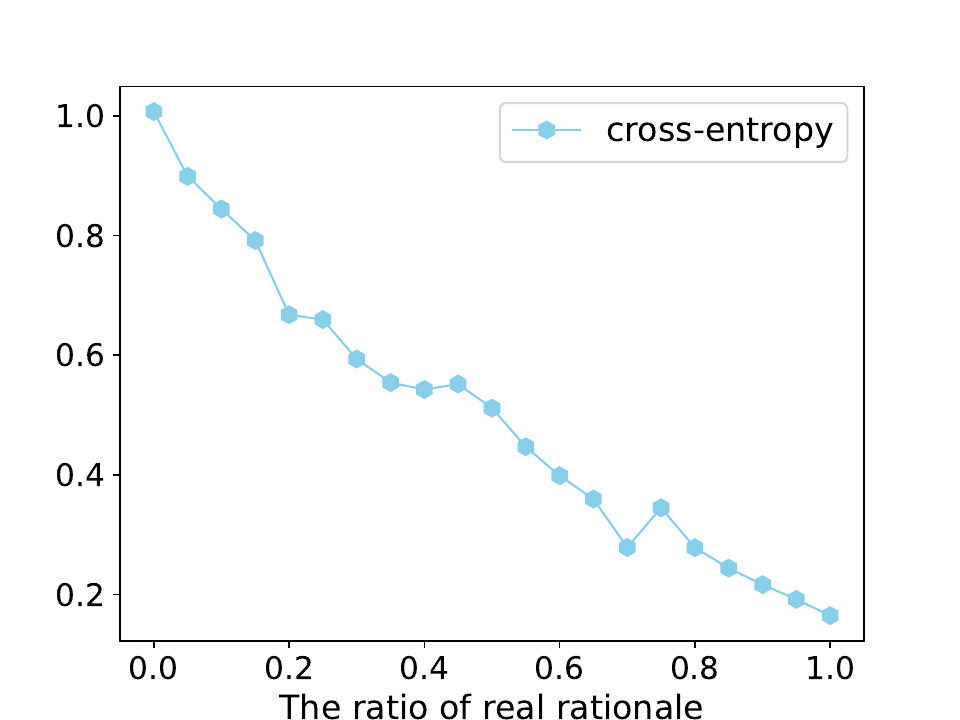}
            \label{fig:rationale_loss hotel1}
        }
        \subfigure[]{
            \includegraphics[width=0.31\columnwidth]{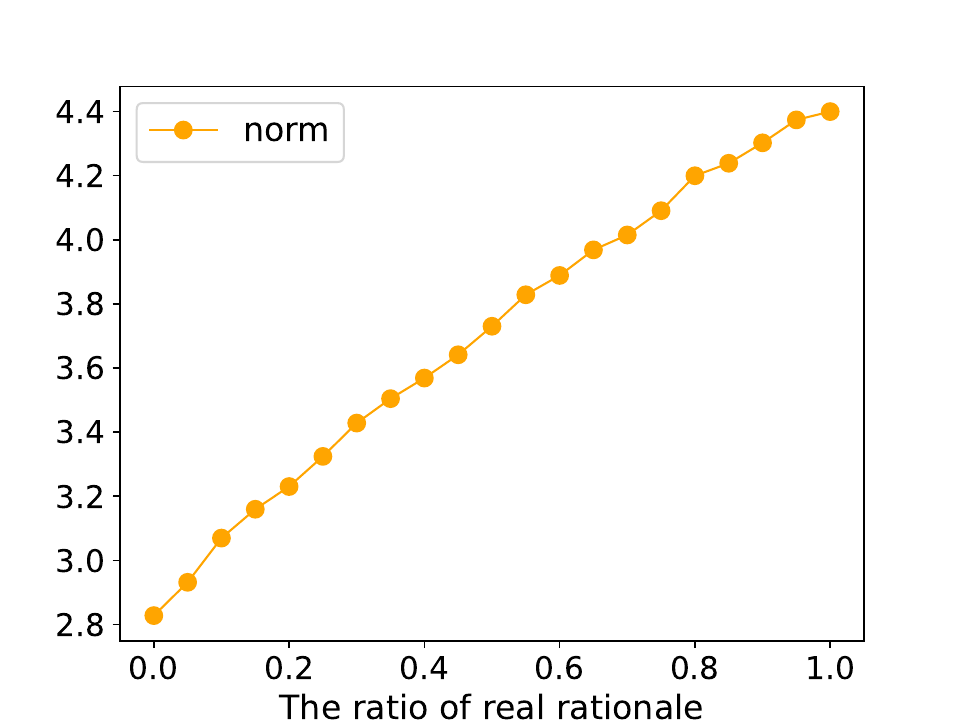}
            \label{fig:rationale_norm hotel1}
        }
        \subfigure[]{
            \includegraphics[width=0.31\columnwidth]{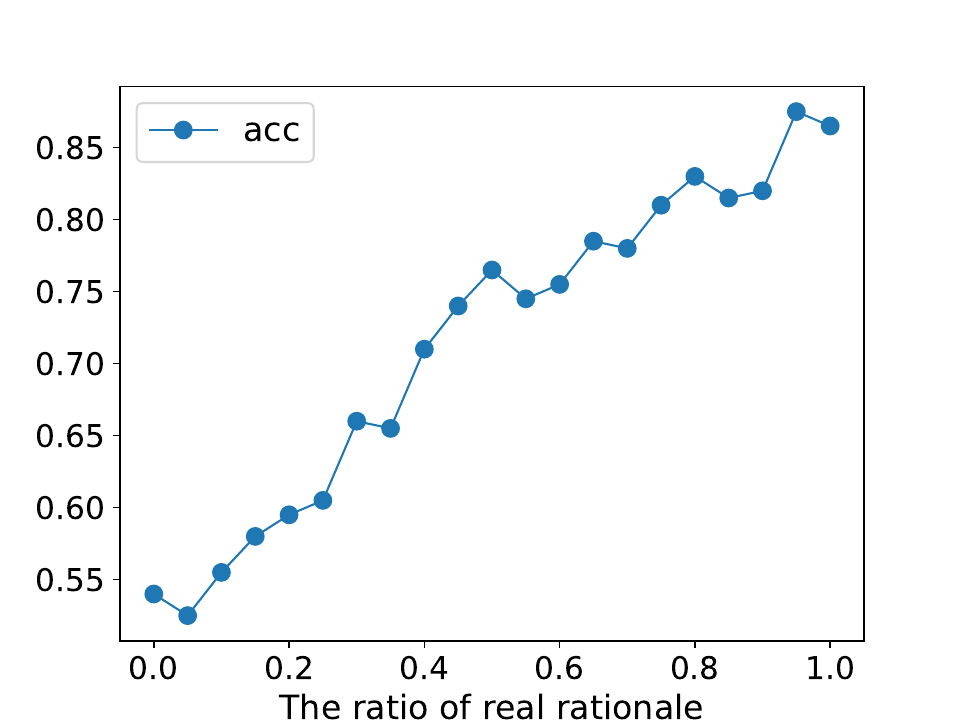}
        \label{fig:rationale_acc1 hotel2}
        }
        \subfigure[]{
            \includegraphics[width=0.31\columnwidth]{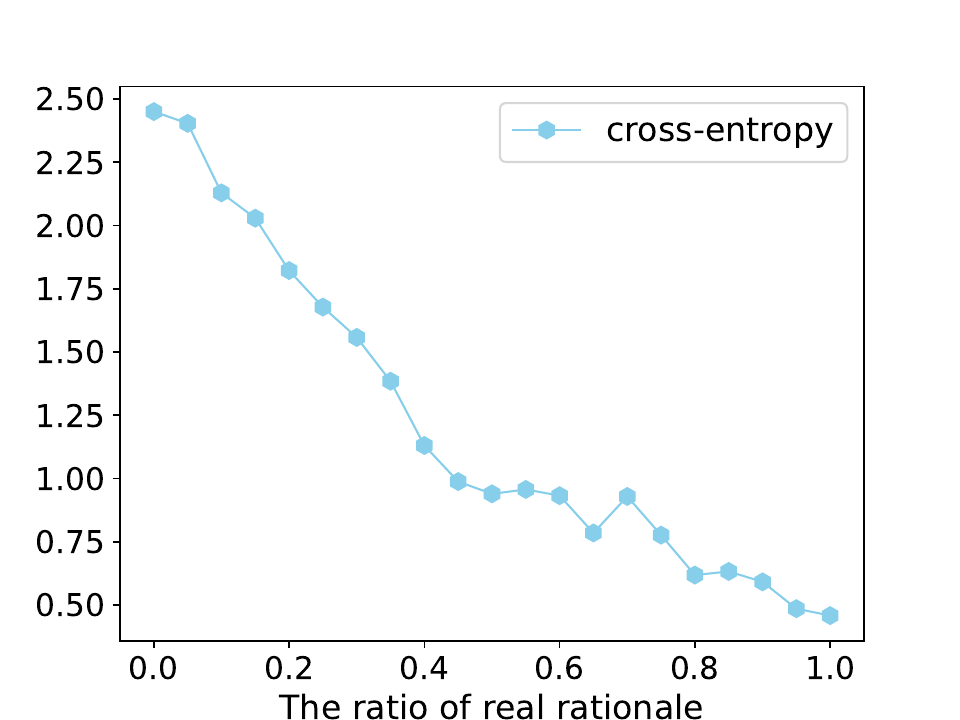}
            \label{fig:rationale_loss hotel2}
        }
        \subfigure[]{
            \includegraphics[width=0.31\columnwidth]{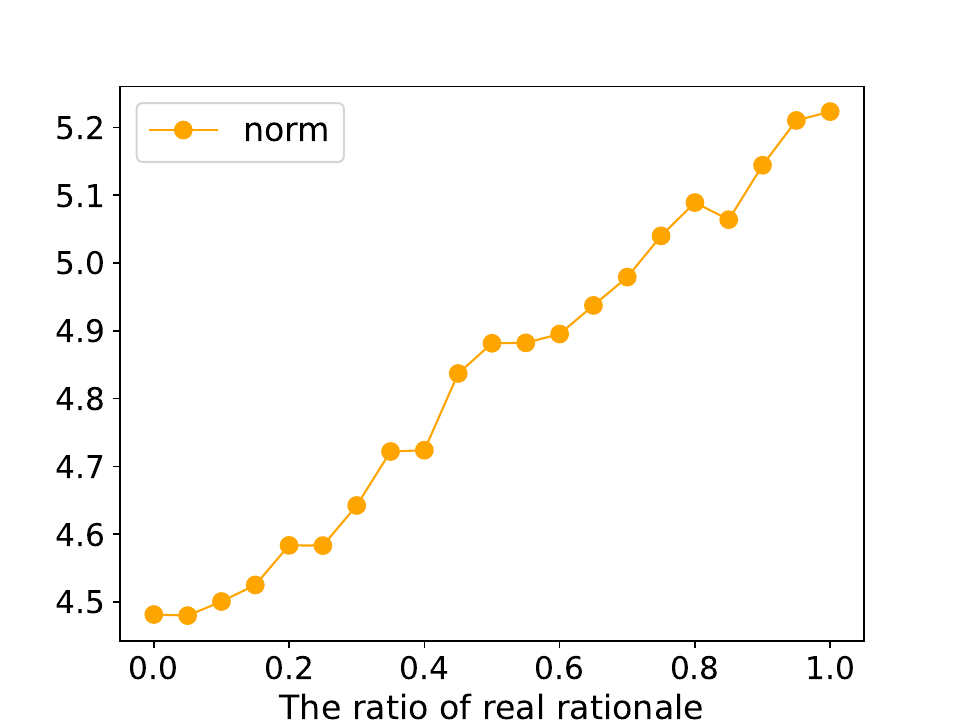}
            \label{fig:rationale_norm hotel2}
        }

    \caption{The prediction accuracy, cross-entropy loss, and the norm of the representation through the neural network vary with the proportion of true rationale components in the rationale candidate input within a trained standard RNP predictor. (a)(b)(c): the Beer-Appearance dataset. (d)(e)(f): the Hotel-Service dataset. (g)(h)(i): the Hotel-Cleanliness dataset.}
    \label{fig:rationale_acc more dataset}
\end{figure*}

Figure \ref{fig:rationale_acc more dataset} shows the results on more datasets.

\subsection{A visualized example of extracted rationales}
\begin{figure}[t]
\centering
    \includegraphics[width=0.99\columnwidth]{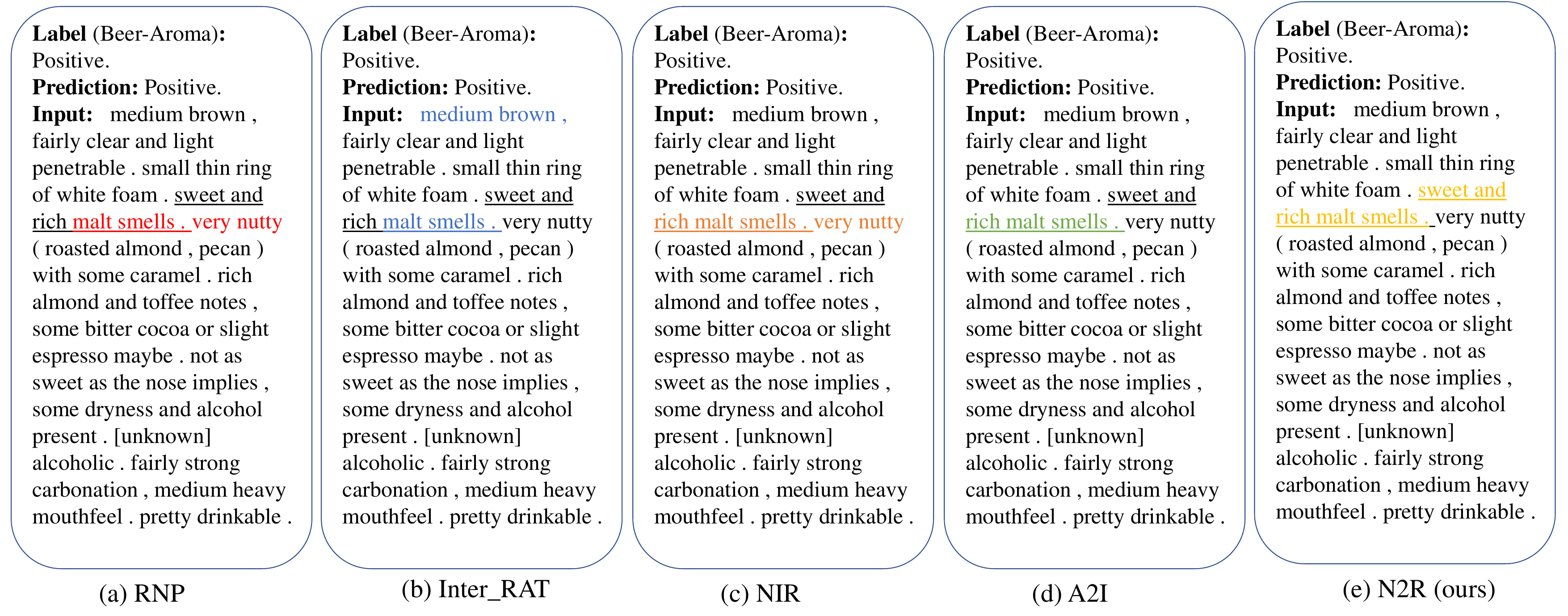}
    \caption{A Visualized example of rationales on the Beer-Appearance dataset. The \underline{underlined} words are human-annotated ground-truth rationales. Model-selected rationales are highlighted with different colors. }
    \label{fig: visualized example}
\end{figure}

Figure \ref{fig: visualized example} shows a visualized example of rationales extracted by different methods on the Beer-Aroma dataset. 

\begin{figure}[t]
\centering
    \includegraphics[width=0.99\columnwidth]{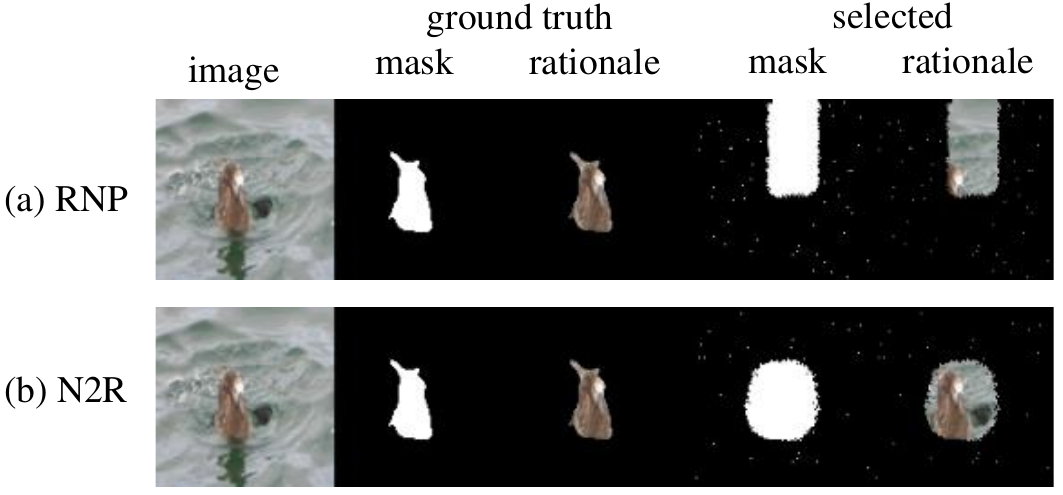}
    \caption{A Visualized example of rationales on the CUB dataset.}
    \label{fig: visualized example of cub}
\end{figure}

Figrue \ref{fig: visualized example of cub} shows a visualized example of rationales extracted by different methods on the CUB (see Appendix \ref{app: cub}) dataset.

\subsection{The results on CUB dataset}\label{app: cub}

\textbf{TL;DR}. The results are in Table \ref{tab: cub}.

The CUB dataset \citep{cub} contains photographs of birds annotated by species, totaling around 11.7k samples. In our experiment, we follow \cite{waterbird} to categorize them into two categories: waterbird (albatross, auklet, cormorant, frigatebird, fulmar, gull, jaeger, kittiwake, pelican, puffin, tern, gadwall, grebe, mallard, merganser, guillemot, or Pacific loon) and landbird (ani, blackbird, bobolink, bunting, cardinal, catbird, Yellow Breasted Chat, Eastern Towhee, Chuck Will Widow, cowbird, Brown Creeper, crow, cuckoo, finch, Northern Flicker, flycatcher, goldfinch, grackle, grosbeak, hummingbird).    \cite{waterbird} annotated each image with the bird's silhouette in order to study the overfitting problem, which can serve as the ground-truth rationale.

The classification task on this dataset is challenging due to severe overfitting problems. Since other baseline rationalization methods  are specifically designed for text data and are not suitable for image data, we only compare with vanilla MMI (i.e., RNP) to validate our method's effectiveness rather than competitiveness. We also compare with a vanilla classifier (ResNet18) to verify the effectiveness of our method in improving classification accuracy.

\textbf{Details and metrics}. For this image classification dataset, the extractor and predictor in Figure \ref{fig:rnp} are different from those used in performing text tasks. The goal of the extractor is to select a portion of the pixels from an image as the rationale, which can be seen as performing a binary classification on each pixel, similar to image segmentation. So, we use an U-Net to be the extractor. And for the predictor, we use a ResNet18. The meaning of F1 is the same as the one used in Table \ref{tab: beer} (the overlap between the extractor-selected pixels and ground truth pixels).
We add an additional metric, IoU (a common metric for image segmentation), to further measure the rationale quality.

\textbf{Results}. The results are shown in Table \ref{tab: cub}. On this challenging image classification dataset, our N2R still outperforms the MMI-based method RNP. As for the improvement in classification accuracy compared to the vanilla classifier ResNet18, one possible reason is that our N2R retains information relevant to classification while removing irrelevant noise (however, the poor performance of RNP may be due to its failure to find comprehensive useful information). This phenomenon is consistent with the findings of \cite{dir}, which suggest that extracting rationales can enhance generalizability to some extent.

We also provide a visualized example for this dataset in Figure \ref{fig: visualized example of cub}. We find that RNP selects only a small part of the bird but a large part of the water (shortcut) to classify this image as "waterbird". While the N2R-selected rationale includes the whole bird.

\begin{table}[]
    \centering

        \caption{The results on the CUB dataset.}
    \vskip 0.1in
    \begin{tabular}{c|c|c|c|c|c|c}
    \hline
    Methods & S & Acc & P &R &F1&IoU \\
    \hline
      RNP   & 15.5 (1.1)& 81.3 (0.7) & 33.3 (2.1) & 37.0 (4.7) & 35.0 (3.2) & 21.3 (2.4) \\
      N2R (ours) & 15.8 (1.1) &\textbf{86.5} (0.6) & \textbf{46.1} (5.4) & \textbf{52.5} (4.8) & \textbf{49.1} (4.9) & \textbf{32.6} (4.2)\\
      Classifier (ResNet18) & n/a & 85.0 (0.8) & n/a& n/a& n/a& n/a\\
      \hline
    \end{tabular}

    \label{tab: cub}
\end{table}

\subsection{Results on MovieReview dataset}
\textbf{TL;DR}. The results are in Table \ref{tab: movie}.

MovieReview \citep{movie} is a text classification dataset with much longer  texts (the average length is 774 words) as compared to the datasets used in Table \ref{tab: beer} and \ref{tab: hotel}. Recently, \cite{eraser} annotated rationales for this dataset so that it can be used for the rationalization task.  

We follow the settings of Inter\_RAT \citep{interventional} to conduct experiments on this dataset. The encoders for extractor and the predictor are both GRUs. And the word embedding is GloVe-100d. The maximum sentence length is set to be 1024. 

We compare with the baselines already implemented by Inter\_RAT and copy the results from it. 
We also compare with a vanilla classifier (GloVe+GRU) to see the classification performance.

The results are shown in Table \ref{tab: movie}. We see that our N2R still outperforms MMI-based methods on this long-text challenging dataset.

\begin{table}[]
    \centering

        \caption{The results on the MovieReview dataset. $``$*": The results of the baselines are copied from Inter\_RAT\citep{interventional} (they did not report the classification accuracy).}
    \vskip 0.1in
    \begin{tabular}{c|l|l|l|l|l}
    \hline
    Methods & S & Acc & P &R &F1 \\
    \hline
      RNP*   & 20.0 &-&35.6 &21.1 &24.1\\
      A2R* &20.0 &-& 48.7 & 31.9 & 34.9\\
      INVRAT* &20.0 &-& 33.9 & 24.3 &28.3\\
      Inter-RAT* &20.0 &- &35.7 & 35.8 & 35.7\\
      N2R & 20.3 (2.2) & \textbf{88.4} (2.1)   & \textbf{45.6} (2.1) &{31.7} (1.6) & \textbf{37.4} (1.9)\\
      Classifier & n/a & 86.4 (2.3) & n/a
      & n/a& n/a\\
      \hline
    \end{tabular}

    \label{tab: movie}
\end{table}

\subsection{Comparison with BERT classifier.}
We compare the classification accuracy with a vanilla classifier implemented with BERT on the Beer-Appearance and Beer-Aroma datasets. The results are shown in Table \ref{tab: bert classifier}. Our N2R gets even better accuracy than the BERT classifier.  One  possible reason is that our N2R retains information relevant to classification while removing irrelevant noise. This phenomenon is consistent with the findings of \cite{dir}, which suggest that extracting rationales can enhance generalizability to some extent.

\begin{table}[t]
    \centering
    \caption{Results with BERT encoder. The dataset is the most widely used Beer-Appearance. $``*"$: results obtained from the paper of CR \citep{cr}. }
    
    \vskip 0.1in
    \resizebox{0.99\columnwidth}{!}{
% \cline{4-8}
\setlength\tabcolsep{4pt}
\begin{tabular}{c c c |c c| c c c|c c| c c c }
\hline
\multicolumn{3}{c|}{\multirow{2}{*}{\diagbox{Methods}{Datasets}}} & \multicolumn{5}{c|}{Beer-Appearance} & \multicolumn{5}{c}{Beer-Aroma} \\
\cline{4-13}
\multicolumn{3}{c|}{} &S& Acc & P & R &\multicolumn{1}{c|}{F1} &S& Acc & P & R &\multicolumn{1}{c}{F1} \\
\hline
         \multicolumn{3}{c|}{RNP*}&10.0 (n/a)& 91.5 (1.7) & 40.0 (1.4) & 20.3 (1.9) & 25.2 (1.7) & 10.0 (n/a) & 84.0 (2.1) & 49.1 (3.2) & 28.7 (2.2) & 32.0 (2.5) \\
         \multicolumn{3}{c|}{A2R*} & 10.0 (n/a)&91.5 (2.2) & 55.0 (0.8)& 25.8 (1.6) & 34.3 (1.4)& 10.0 (n/a) & 85.5 (1.9)& 61.3 (2.8) & 34.8 (3.1) & 41.2 (3.3) \\
         \multicolumn{3}{c|}{INVRAT*} &10.0 (n/a) & 91.0 (3.1)& 56.4 (2.5) & 27.3 (1.2) & 36.7 (2.1)& 10.0 (n/a) & 90.0 (3.0) & 49.6 (3.1) & 27.5 (1.9) & 33.2 (2.6)\\
         \multicolumn{3}{c|}{CR*}&10.0 (n/a)& 92.4 (1.7)& 59.7 (1.9)&31.6 (1.6)&39.0 (1.5)& 10.0 (n/a) & 86.5 (2.1) & 68.0 (2.9) & 42.0 (3.0) & 49.1 (2.8)\\
         \multicolumn{3}{c|}{N2R (ours)}& 10.8 (0.3)&\textbf{93.5} (1.8) & \textbf{79.7} (4.1) & \textbf{36.3} (1.8) & \textbf{49.9} (2.5)& 10.0 (0.1) & 91.0 (3.6) & \textbf{74.3} (5.8) & \textbf{47.0} (3.7) & \textbf{57.6} (4.5)\\
         \multicolumn{3}{c|}{Classifier}& n/a & 93.0 (2.6) &n/a&n/a&n/a&n/a &\textbf{91.6} (3.1) &n/a&n/a&n/a  \\
         \hline

    \end{tabular}
   }
   
    \label{tab: bert classifier}

\end{table}

\subsection{The convergence speed}
\begin{figure*}[t]
    % \flushleft
      \subfigure[]{
            \includegraphics[width=0.46\columnwidth]{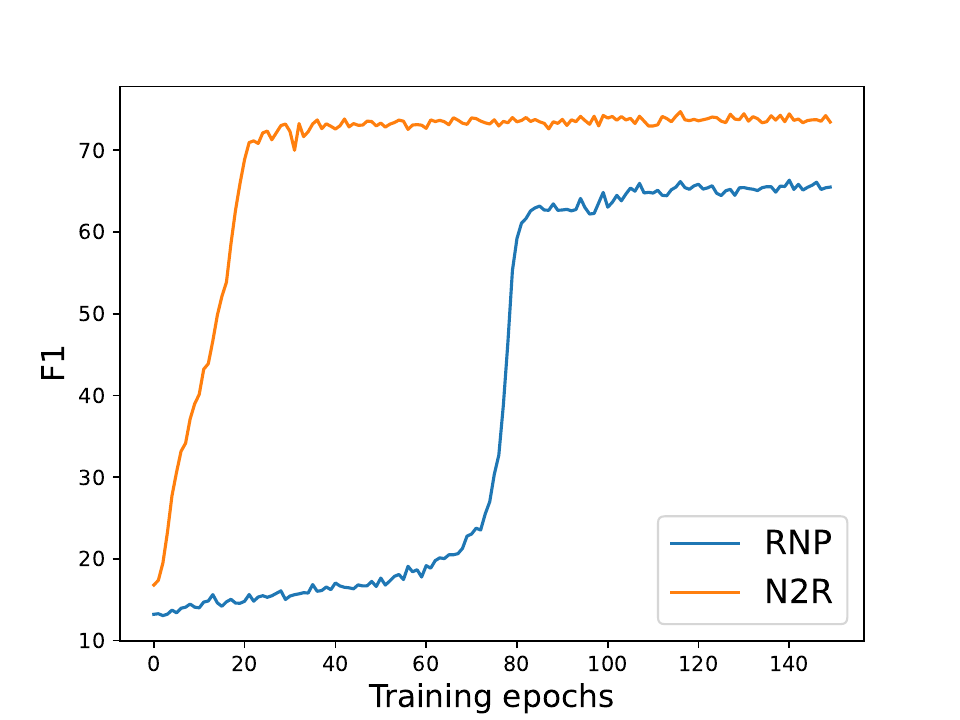}
        \label{fig: convergence beer0}
        }
        \subfigure[]{
            \includegraphics[width=0.46\columnwidth]{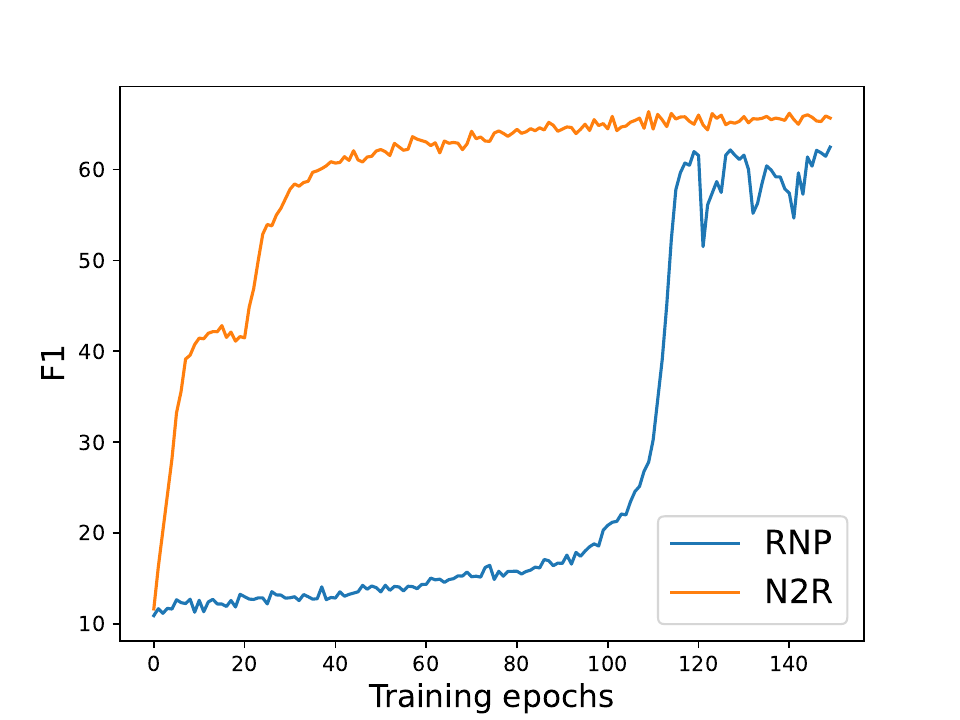}
            \label{fig: convergence beer1}
        }

    \caption{The comparison of convergence speed between RNP and N2R. (a) The Beer-Appearance dataset. (b) The Beer-Aroma dataset.}
    \label{fig: convergence}
    % \vspace{-10pt}
\end{figure*}

Figure \ref{fig: convergence} shows a comparison of convergence speed between RNP and N2R. The batchsize is 128 and the learning rate is 0.0001. We see that N2R converges much faster than RNP, which means that we can train N2R with fewer steps and thus saving the computational costs.

\subsection{Theoretical Support for Using Norms to Distinguish Between Rationales}\label{app: theoretical support for the low norm}

\begin{lemma}[\citep{highdimvec}]\label{lem: high-dim vec}
Let $U$ and $V$ be two random points on a $p$-dimensional unit hypersphere $\mathbb{R}^p$, and $O$ is the origin. Let $\Theta$ be the angle between vector $\overrightarrow{OU}$ and vector $\overrightarrow{OV}$, then 
\begin{equation}
    \Pr(|\Theta-\frac{\pi}{2}|\geq \epsilon)\leq K\sqrt{p}(\cos\epsilon)^{p-2},
\end{equation}
for all $p\geq 2$ and $\epsilon \in (0,\pi/2)$, where $K$ is a universal constant.
\end{lemma}

Lemma \ref{lem: high-dim vec} is the Proposition 5 of \citep{highdimvec}. And it tells us that $``$all high-dimensional random vectors are almost always nearly orthogonal to each other" \citep{highdimvec}. As the dimension $p$ increases, $\Theta$ gradually converges to $\pi/2$.

If two vectors are orthogonal to each other, then their dot product will be zero. 

We consider a simple case. Consider a FFN layer without the bias term. And its weight matrix is $W\in \mathbb{R}^{m\times n}$, and the input is $X\in \mathbb{R}^m$. The output of the FFN layer is $XW\in \mathbb{R}^n$. Each dimension in $XW$ is the dot product of $X$ and a column vector of $W$.The column vectors of $W$ represent some directions the model can handle. For a well-trained FFN, the directions are determined by its training data and represent the information of it learns. 

We use the lower case $x$ to denote a specific input.

Consider an input $x_1$ does not contain any information that has been learned by $W$. For example, $W$ is part of a classifier trained to distinguish between cell phones and water cups, and $x_1$ is a photo of a grassy. Although $x_1$ is not truly random noise, since $W$ has never been trained to recognize grassland, $x_1$ acts like random noise to $W$. So it is  likely that $x_1$ is orthogonal to those column vectors in $W$. So, every dimension of $x_1W$ will approach zero and thus $||x_1W||_2$ will also approach zero.

If an input $x_2$ is full of the information learned by $W$, then it probably matches the directions of $W$. And thus $||x_2W||_2$ will be high.

Then, what would be the intermediate state of $x_1$ and $x_2$? \textbf{We use $V\in \mathbb{R}^m$ to denote an arbitrary column vector in $W$.} We assume an input $x_3$. The first $k$ dimensions of $x_3$ (denoted as $x_3^{1:k}$) contain useful information, and the last $n-k$ dimensions represent uninformative noise. We have that 
\begin{equation}
    x_3\cdot V= x_3^{1:k}\cdot V^{1:k}+x_3^{k+1:n}\cdot V^{k+1:n}
\end{equation}
$x_3^{k+1:n}$ consists of uninformative noise, so we have $x_3^{k+1:n}\cdot V^{k+1:n}\approx 0$. Thus $x_3\cdot V= x_3^{1:k}\cdot V^{1:k}$.

We assume that the useful information is uniformly distributed in each dimension of $X$ and $V$. For an input $x$, we denote $x^i$ as the $i$-th dimension of it. Formally:
\begin{assumption}\label{ass: different dim is same}
    If an input $x$ contains $k$-dimensional useful information (for simplicity, we always assume that the useful information is in the first $k$ dimensions), we assume that the informativeness between different dimensions is the same:
    \begin{equation}
        x^i\cdot V^i=x^j\cdot V^j, \ \forall i\leq k, j\leq k.
    \end{equation}
\end{assumption}

With this assumption in place, we can quantitatively analyze why our method is not affected by the problem of diminishing marginal returns.

Consider that the proportion of the gold rationale components in an extractor-selected rationale candidate $x_a$ is $\frac{k_a}{n}$ (i.e., there are $k_a$ of $n$ dimensions in $x_a$ represent useful information). When we add more gold rationale components to it to make the proportion be $k_a'$ and the new rationale candidate be $x_a'$. We will have that 
\begin{equation}
    \frac{||x_aW||_2}{||x_a'W||_2}=\frac{k_a}{k_a'}.
\end{equation}

\textbf{Proof}. 
\begin{equation}
\begin{aligned}
    x_aW&=[x_a\cdot V_1,\cdots , x_a\cdot V_n]\\
    &=[x_a^{1:k_a}\cdot V_1^{1:k_a},\cdots ,x_a^{1:k_a}\cdot V_n^{1:k_a}]\\
    &=k_a[x_a^{1}\cdot V_1^{1},\cdots ,x_a^{1}\cdot V_n^{1}],
\end{aligned}
\end{equation}
where the third equation is from Assumption \ref{ass: different dim is same}. 

Similarly, we have  
\begin{equation}
\begin{aligned}
    x_a'W&=k_a'[x_a'^{1}\cdot V_1^{1},\cdots ,x_a'^{1}\cdot V_n^{1}],
\end{aligned}
\end{equation}
$x_a'$ is got from replacing the uninformative part of $x_a$ (i.e., the last a few dimensions) with useful information, so we have $x_a'^{1}=x_a^{1}$. Thus we have $\frac{x_aW}{x_a'W}=\frac{k_a}{k_a'}$ and $\frac{||x_aW||_2}{||x_a'W||_2}=\frac{k_a}{k_a'}$.

\textbf{Conclusion}. Ideally, the norm metric we designed should increase approximately linearly as the proportion of gold rationale in the rationale candidates grows, thus avoiding the problem of diminishing marginal returns like MMI-based methods.

Although we made an assumption that may not hold in reality, it was made to facilitate a better quantitative analysis, and the conclusions drawn at least have a qualitative trend. The trends in Figure \ref{fig:rationale_acc more dataset}(c)(f)(i) also verify this trend.

\subsection{A toy example for better intuitive understanding of our method}\label{app: toy example}
Consider a network consists of only a linear layer (without bias) and has a low-rank weight matrix $M$ that, after elementary row transformations, takes the following form:
\[
\begin{bmatrix}
1 & 1 & 1 \\
0 & 1 & 1 \\
0 & 0 & 1\\
0 & 0 & 0

\end{bmatrix}
\]

$A$ and $B$ represent inputs with and without the model's learned information respectively. After undergoing corresponding transformations (existing research indicates that after performing certain elementary transformations on the weight matrix, a corresponding transformation can be found for the inputs, resulting in identical outputs \citep{gitrebis}), they are likely to take the forms like: $A=[1,0,0,0]$, $B=[0,0,0,1]$. The informative input $A$ usually matches more with the non-zero parts of $M$. And $||AM||_2=\sqrt{3}>||BM||_2=0$. (The column vectors of the weight matrix represent certain directions in high-dimensional space (determined by the learned information), but random noise $B$ generally does not fall along these directions as it does not contain the corresponding information.)

\end{document}